\documentclass{article}

\usepackage{microtype}
\usepackage{graphicx}
\usepackage{subfigure}
\usepackage{booktabs} 

\usepackage{hyperref}

\usepackage[accepted]{icml2021}

\icmltitlerunning{ARMS: Antithetic-REINFORCE-Multi-Sample Gradient for Binary Variables}

\usepackage{bm}
\usepackage{bbm}
\usepackage{amsmath}
\usepackage{amssymb}
\usepackage{amsfonts}
\usepackage{amsthm}
\usepackage{multirow}
\usepackage{multicol}
\usepackage{float}
\usepackage{array}

\newtheorem{theorem}{Theorem}

\newcommand{\Var}[0]{\text{Var}}
\newcommand{\E}[0]{\mathbb{E}}

\newcommand{\iid}{\overset{iid}{\sim}}
\newcommand{\g}[1]{g_{\text{#1}}}

\begin{document}

\twocolumn[
\icmltitle{ARMS: Antithetic-REINFORCE-Multi-Sample Gradient for Binary Variables}

\begin{icmlauthorlist}
\icmlauthor{Alek Dimitriev}{ut}
\icmlauthor{Mingyuan Zhou}{ut}
\end{icmlauthorlist}
\icmlaffiliation{ut}{McCombs School of Business, University of Texas at Austin, Austin, Texas, USA}
\icmlcorrespondingauthor{Alek Dimitriev}{alekdimi@utexas.edu}

\icmlkeywords{reinforce, score function, antithetic, gradient, estimator, binary, latent, variables, variational, autoencoder, disarm}

\vskip 0.3in
]

\printAffiliationsAndNotice{}
    
\begin{abstract}
Estimating the gradients for binary variables is a task that arises frequently in various domains, such as training discrete latent variable models. What has been commonly used is a REINFORCE based Monte Carlo estimation method that uses either independent samples or pairs of negatively correlated samples. To better utilize more than two samples, we propose ARMS, an Antithetic REINFORCE-based Multi-Sample gradient estimator. ARMS uses a copula to generate any number of mutually antithetic samples. It is unbiased, has low variance, and generalizes both DisARM, which we show to be ARMS with two samples, and the leave-one-out REINFORCE (LOORF) estimator, which is ARMS with uncorrelated samples. We evaluate ARMS on several datasets for training generative models, and our experimental results show that it outperforms competing methods. We also develop a version of ARMS for optimizing the multi-sample variational bound, and show that it outperforms both VIMCO and DisARM. The code is publicly available\footnote{https://github.com/alekdimi/arms}.
\end{abstract}

\section{Introduction}
\label{introduction}

At the heart of many optimization problems is optimizing an expectation of a function with respect to the parameters of the distribution, such as $\mathcal{E}(\bm \phi) = \E_{\bm b}[f(\bm b)]$, where $\bm b \sim p_{\bm \phi}(\bm b)$. There are several approaches to estimate the gradient $\nabla_{\bm \phi} \E_{\bm b}[f(\bm b)]$. Most commonly used is the reparameterization gradient~\cite{kingma2013auto, rezende2014stochastic}. If $f$ is differentiable, and $b$ can be expressed as a differentiable deterministic transformation $b = \mathcal{T}_{\bm \phi}(\bm \epsilon)$ of a variable $\bm \epsilon \sim p(\bm \epsilon)$ not dependent on $\bm \phi$, then:
\begin{align*}
    \nabla_{\bm \phi} \mathcal{E}(\bm \phi) = \nabla_{\bm \phi} \E_{\bm \epsilon} \Big[  f\big( \mathcal{T}_{\bm \phi}(\bm \epsilon) \big) \Big] = \E_{\bm \epsilon} \Big[ \nabla_{\bm \phi} f\big( \mathcal{T}_{\bm \phi}(\bm \epsilon) \big) \Big].
\end{align*}
If $\bm b$ is discrete, however, $\mathcal{T}_{\bm \phi}$ is not differentiable, and $f$ is not always differentiable, $e.g.$, if it corresponds to the reward function in reinforcement learning. This has inspired work into replacing $\mathcal{T}_{\bm \phi}$ (and $f$) with a continuous relaxation for the discrete case, using the Gumbel-Softmax trick~\cite{maddison2017concrete, jang2017categorical}. Some are biased~\cite{bengio2013estimating, lorberbom2018direct}, and others, such as REBAR~\cite{tucker2017rebar}, and RELAX~\cite{grathwohl2018relax}, contain a debiasing term. However, a relaxation could require much more computation, $e.g.$, in best subset selection~\cite{yin2020probabilistic} where $f(\bm b)$ is the loss function, and $\bm b$ is a (sparse) binary vector that selects a subset of features on which to train the model. Evaluating a relaxed version of $f$ for any input would train the model with all features at considerable computational cost if the number of features is large.

A second, less well-known approach, is the measure valued gradient~\cite{rosca2019measure, mohamed2020monte}. If possible, it expresses the $d^{\text{th}}$ element of the gradient as a difference of two measures: $\nabla_{\bm \phi^d} p_{\bm \phi}(\bm b) = c^d( p^d_+(\bm b)  - p^d_-(\bm b) )$
This allows us to estimate $\nabla_{\bm \phi^d} \mathcal{E}(\bm \phi)$ using as few as two $f$ evaluations per dimension:
\begin{equation*}
    \nabla_{\bm \phi^d} \mathcal{E}(\bm \phi) = c^d \Big( \E_{\bm b \sim p^d_+} \big[ f(\bm b) \big] - \E_{\bm b' \sim p^d_-} \big[ f(\bm b) \big] \Big)
,\end{equation*}
but needs to be done for each dimension separately, which is not scalable for a large number of dimensions, requiring twice as many function evaluations.

The most widely applicable approach is the score function estimator~\cite{glynn1990likelihood, fu2006gradient}, also known as REINFORCE~\cite{williams1992simple}. Using the log derivative trick $  \nabla_{\bm \phi} p(\bm b) = p(\bm b) \nabla_{\bm \phi} \ln p(\bm b) $, it expresses the gradient as an expectation with respect to the original distribution:
\begin{align*}
   \nabla_{\bm \phi} \mathcal{E}(\bm \phi) &= \nabla_{\bm \phi} \E_{\bm b} \big[ f(\bm b) \big] = \E_{\bm b} \big[ f(\bm b) \nabla_{\bm \phi} \ln p(\bm b) \big]
,\end{align*}
which results in a simple form, only requiring the ability to sample from the original distribution. Unlike the reparameterization trick, it is directly applicable to discrete variables, but usually suffers from high variance in the form presented above. Recent work has focused on various mechanisms for variance reduction, applied either to general variational inference tasks~\cite{paisley2012variational, ranganath2014black, ruiz2016generalized, kucukelbir2017automatic} or focusing on discrete \cite{mnih2014neural, gu2015muprop} cases. Leveraging the specific structure of the distribution $p_{\bm \phi}(\bm b)$, while preserving its unbiasedness, has also proven useful~\cite{titsias2015local}. 

Antithetic variates~\cite{owen2013monte}, which are negatively correlated samples, have recently gained popularity as a variance reduction method~\cite{wu2019differentiable, ren2019adaptive}. The first estimator to use them for binary variables is the augment-REINFORCE-merge (ARM) estimator of \citet{yin2019arm}, which reparameterizes any binary variable using a uniform distribution that has an antithetic pair. Coupling the estimates for both uniform variables results in much lower variance.  Another approach is the unbiased uniform gradient (U2G) estimator of \citet{yin2020probabilistic}, which is independently discovered  by~\citet{dong2020disarm} and referred to as DisARM. These two equivalent estimators, which are derived using different methods, improve upon ARM by marginalizing out the continuous reparameterization.

Our main contribution is building upon the idea of using antithetic pairs and extending it to $n$ samples that are jointly antithetic, resulting in a novel unbiased, low variance gradient estimator for binary variables, which we denote as the Antithetic REINFORCE Multi Sample (ARMS) gradient estimator. We also develop a version of ARMS for optimizing a multi-sample bound~\cite{burda2016importance}, which outperforms both VIMCO~\cite{mnih2016variational} and DisARM.  

\section{Background}
\label{background}

Let $\bm b = (b^1, ..., b^m)$ be a vector of $m$ independent Bernoulli variables with $b^d \sim p_{ \phi^d}(b^d) = \text{Bern}(\sigma(\phi^d))$, where $\sigma(x) = 1 / (1 + e^{-x})$ is the sigmoid function and $\bm \phi = (\phi^1, ..., \phi^m)$.  This paper is focused on optimizing a factorized Bernoulli expectation with respect to the logits $\bm \phi$:
\begin{equation}
    \label{eq:factorized}
    \mathcal{E}(\bm \phi) = \E_{\bm b} [ f(\bm b)], \quad \bm b \sim p_{\bm \phi}(\bm b) = \prod_{d=1}^m p_{\bm \phi^d} (b^d)
.\end{equation}
Throughout the paper, we use superscripts to refer to dimensions of the Bernoulli vector and subscripts to refer to particular samples. This form arises, for example, in variational inference, where a mean-field approximation is commonly used for the latent space of a variational autoencoder. Note that $f$ could depend on $\bm \phi$, but there are no difficulties in estimating $\E_{\bm b}[\nabla_{\bm \phi} f_{\bm \phi}(\bm b)]$, so without loss of generality, we omit this gradient term for notational simplicity.

\subsection{LOORF}
The Leave One Out REINFORCE (LOORF) estimator \cite{salimans2014using, kool2019buy} is a simple REINFORCE baseline that utilizes $n$ independent samples. If $\bm b_i \iid p_{\bm \phi}(\bm b)$, $i = 1, ..., n$, then an unbiased estimate of the gradient of $\nabla_{\phi} \mathcal{E}(\bm \phi)$ is:
\begin{align}
    \label{eq:baseline}
    &\g{LOORF}(\bm b) =  \resizebox{.74\columnwidth}{!}{$ \displaystyle\frac{1}{n} \sum_{i=1}^n  \bigg( f(\bm b_i) - \frac{1}{n-1} \sum_{j \neq i} f(\bm b_j) \bigg) \nabla_{\bm \phi} \ln p_{\bm \phi}(\bm b_i)$} \nonumber \\
    &\;= \frac{1}{n - 1} \sum_{i=1}^n \bigg( f(\bm b_i) - \frac{1}{n} \sum_{i=1}^n f(b_j) \bigg) \nabla_{\bm \phi} \ln p_\phi(\bm b_i)     
.\end{align}
The latter form is simpler to implement because we can precompute the average once and subtract it for each sample. The proofs of the unbiasedness and equivalence of the two forms are known~\cite{kool2019buy}, but provided for completeness in Appendix~\ref{appendix:proofs}, along with all subsequent proofs.

\subsection{ARM}
The first binary estimator to use antithetic samples is ARM~\cite{yin2019arm}, which accomplishes this by reparameterizing the gradient with the antithetic pair $(u, 1 - u)$. We review the univariate case below. Using the observation that $\int_0^p (1 - 2u) du = p(1 - p)$, they rewrite the analytical univariate gradient into the following form, which they denote as the augment-REINFORCE (AR) estimator:
\begin{align*}
    &\nabla_\phi \mathcal{E}(\phi) = \big( f(1) - f(0) \big) p(1 - p) \\
    &= \E_{u} [ f \big( \mathbbm{1}_{u < p}(1 - 2u \big)] = \E_{u}[ \g{AR}(u)], \;u \sim \text{Unif}(0, 1),
\end{align*}
where $\mathbbm{1}_a$ is the indicator function having the value 1 if $a$ is true and 0 otherwise. Since both $u, 1 - u \sim \text{Unif}(0, 1)$, they average $\g{AR}(u)$ and $\g{AR}(1 - u)$, which is still unbiased, and arrive at the univariate ARM estimator:
\begin{align*}
&\E_u [ \g{ARM}(u)] = \E_u [ \g{AR}(u) + \g{AR}(1 - u)] = \\ 
&= \frac{1}{2} \E_{u} \Big[ f(\mathbbm{1}_{u < p})(1 - 2u) + f(\mathbbm{1}_{(1 - u) < p}) (2u - 1) \Big] \\
    &= \E_u \left[ \Big( f(\mathbbm{1}_{u < p}) - f(\mathbbm{1}_{u > 1 - p}) \Big) \left( \frac{1}{2} - u \right) \right].
\end{align*}

\subsection{DisARM}
\cite{dong2020disarm} observe that although ARM reduces variance by using an antithetic pair, it also increases the variance by using a continuous reparameterization. This leads them to condition on the discretized pair $(b, b') = (\mathbbm{1}_{u < p}, \mathbbm{1}_{1 - u < p})$ and integrate out $u$ to obtain DisARM:
\begin{align*}
    \g{DisARM}(b, b') &= \E_{u \sim p(u | b, b')} [\g{ARM}(u)] \\
    &= \tfrac{1}{2} ( f(b) - f(b'))(b - b') \max(p, 1 - p).
\end{align*}
This is still unbiased due to the law of total expectation: $E_u[\g{ARM}(u)] = \E_{b, b'} \big[ \E_{u | b, b'}[\g{ARM}(u)]\big]$, and also has lower or equal variance due to the law of total variance:
\begin{align*}
    &\resizebox{.99\columnwidth}{!}{$ \displaystyle \Var(\g{ARM}(u)) = \E \big[\Var_{u | b, b'}(\g{ARM}(u) \big] + \Var \big(\E_{u | b, b'}[\g{ARM}(u)] \big) $} \\
    &\quad\quad\quad\quad\quad\; \resizebox{.75\columnwidth}{!}{$ \displaystyle \geq \Var \big( \E_{u | b, b'}[\g{ARM}(u)] \big) = \Var(\g{DisARM}(b, b')).  $}
\end{align*}

\subsection{Copulas}

A multivariate distribution $\bm u = (u_1, ..., u_n) \sim \mathcal{C}_n$, with marginal distributions $u_i \sim \text{Unif}(0, 1)~ \forall i$ is called a copula. See~\cite{trivedi2007copula} for a more detailed overview. We are interested in the case where there is strong negative dependence between each pair $(u_i, u_j)$. The lower Fréchet–Hoeffding copula bound puts a limit on the negative dependence:
\begin{equation*}
    C(u_1, \dots, u_n) \geq \max \Big\{ 1 - n + \sum_{i=1}^n u_i, 0 \Big\}.
\end{equation*}
The lower bound is a CDF only when $n=2$, and $C(u_1, u_2) = \max(u_1 + u_2 - 1, 0)$ corresponds to the pair $(u, 1 - u), u \sim \text{Unif}(0, 1)$. However for larger $n$ this bound is not sharp~\cite{trivedi2007copula}, so there are no perfectly negatively dependent copulas in higher dimensions. To generate a copula sample, the probability integral transform is frequently used:
\begin{align}
    \label{eq:cdftransform}
    P(F_x(X) &< u) = P(F^{-1}_x(F_x(X)) < F^{-1}_x(u)) \nonumber \\
    &= P(X < F^{-1}_x(u)) = F_x(F^{-1}_x(u)) = u,
\end{align}
where we assumed $F_x$ is a bijection to simplify the proof. If we know each marginal CDF $F_{x_i}(x)$ of a multivariate sample $\bm x = (x_1, ..., x_n)$, then a copula sample is $\big(F_{x_1}(x_1), ..., F_{x_n}(x_n) \big)$. 

\section{ARMS}

In the univariate case with $n=2$ samples, antithetic pair estimators like (Dis)ARM have lower variance than LOORF, but we observed that even on a toy example, they are outperformed by LOORF as the number of samples rises. This motivates us to find a way to generate $n$ jointly antithetic samples, and use them to construct an unbiased estimator. For clarity, we first derive the univariate case of our jointly antithetic estimator, which is obtained by starting with two independent samples, extending this to two arbitrarily correlated Bernoulli samples, and then generalizing to $n$ samples.

For two samples, LOORF has the following equivalent simple form, which we denote as the Product of Differences (PoD) estimator:
\begin{equation}
    \label{eq:pod}
    \resizebox{.9\columnwidth}{!}{$ \displaystyle \g{PoD}(b, b') = \frac{1}{2} \Big( f(b) - f(b') \Big) \Big( \nabla_{\phi} \ln p_{\phi}(b) - \nabla_{\phi} \ln p_{\phi}(b') \Big). $}
\end{equation}
Unless $b$ and $b'$ are independent, this estimator will be biased. However, when $b$ and $b'$ are both Bernoulli variables, we can obtain an unbiased estimator for any bivariate distribution, by using a multiplicative debiasing term. Let $(b, b') \sim \mathcal{B}_2(p)$ denote a sample from a bivariate Bernoulli distribution with marginals $b, b' \sim \text{Bern}(p)$ and $\text{Corr}(b, b') = \rho$. In this case, $\g{PoD}(b, b') = (f(b) - f(b')(b - b')/2$, whose expectation is expressed as:
\begin{equation*}
    \E_{b, b'} \big[ \g{PoD}(b, b') \big] = \big( f(1) - f(0) \big) P(b = 1, b' = 0),
\end{equation*}
with the analytical gradient being $\nabla_\phi \mathcal{E}(\phi) = (f(1) - f(0))p(1 - p)$. Therefore, multiplying the above expression with $p(1 - p)/P(b = 1, b' = 0)$ results in an unbiased estimator for any dependence structure:
\begin{align*}
    \E_{b, b'} &\Bigg[ \frac{p(1 - p)}{P(b = 1, b'= 0)} \g{PoD}(b, b') \Bigg] \\
    &\;= \frac{p(1 - p)}{P(b = 1, b'= 0)} \big( f(1) - f(0) \big) P(b = 1, b' = 0) \\
    &\;= \big( f(1) - f(0) \big) p(1 - p) = \nabla_\phi \mathcal{E}(\phi).
\end{align*}
Lastly, note that:
\begin{align*}
    P(b &= 1, b' = 0) = p - P(b = 1, b' = 1) \\
    &= 2(p - p(1 - p)\rho - p^2) = p(1 - p)(1 - \rho)
,\end{align*}
which simplifies the multiplicative term to $1 / (1 - \rho)$. We summarize the derivation with the following theorem, denoting it the Antithetic-REINFORCE-Two-Sample (ARTS) gradient estimator.
\begin{theorem}
    \label{thm:arts}
    Let $(b, b') \sim \mathcal{B}_2(\sigma(\phi))$ be a sample from a bivariate Bernoulli distribution with marginal distributions $b, b' \sim \text{\emph{Bern}}(\sigma(\phi))$ and correlation $\rho = \text{\emph{Corr}}(b, b')$. An unbiased estimator of $\nabla_\phi \E_b [f(b)]$ is:
    \begin{equation}
        \g{\emph{ARTS}}(b, b', \rho) = \Big( f(b) - f(b') \Big) \frac{b - b'}{2(1 - \rho)}.
    \end{equation}
\end{theorem}

Using Theorem~\ref{thm:arts}, we can easily derive DisARM/U2G, since it is just ARTS with a specific correlation. The antithetic pair used is:
\begin{equation*}
    (b, b') = (\mathbbm{1}_{u < p}, \mathbbm{1}_{(1 - u) < p}), \; u \sim \text{Unif}(0, 1),
\end{equation*}
in which case:
\begin{align*}
    P(b = 1, &b' = 0) = P(u < p, 1 - u > p) \\
    &= P(u < \min(p, 1 - p)) = \min(p, 1 - p).
\end{align*}
The multiplicative term is $p(1 - p) / \min(p, 1 - p) = \max(p, 1 - p)$, which results in the DisARM/U2G estimator:
\begin{align*}
    \g{DisARM}(b, b') = \frac{1}{2}\big( f(b) - f(b') \big) (b - b') \max(p, 1 - p).
\end{align*}
The next theorem shows that in the two sample case, the bivariate distribution with the lowest gradient variance uses the same pair of antithetic variables. The proof is given in Appendix~\ref{appendix:proofs}.

\begin{theorem}
    \label{thm:optimality}
    Let $b, b' \iid \text{Bern}(p)$, and $(\tilde b, \tilde b') \sim \mathcal{B}(p)$. Then:
    \begin{equation*}
        \rho < 0 \implies \Var(\g{\emph{ARTS}}(\tilde b, \tilde b', \rho)) < \Var(\g{\emph{PoD}}(b, b')).
    \end{equation*}
    Furthermore, $\Var(\g{\emph{ARTS}}(\tilde b, \tilde b', \rho))$ is a decreasing function of $\rho$ and achieves its minimum at: 
    \begin{equation*}
        \rho_{\text{min}} = -\min \bigg( \frac{p}{1 - p} , \frac{1 - p}{p} \bigg),
    \end{equation*}
    with a corresponding debiasing term $1 / (1 - \rho_{\text{min}}) = \text{max}(p, 1 - p)$. If $(b, b') = (\mathbbm{1}_{u < p}, \mathbbm{1}_{1 - u < p})$, where $u \sim \text{\emph{Unif}}(0, 1)$, then $\text{Corr}(b, b') = \rho_{\text{min}}$. 
\end{theorem}

We now extend ARTS to $n$ samples. Let $\bm b = (b_1, ..., b_n) \sim \mathcal{B}_n(p)$ with marginals $b_i \sim \text{Bern}(p)$ and correlations $\rho_{ij} = \text{Corr}(b_i, b_j)$. If we compute $\g{ARTS}(b_i, b_j, \rho_{ij})$ for all $\binom{n}{2}$ pairs and take the average, we obtain an unbiased estimator because of the linearity of expectations:
\begin{align*}
    &\E_{b_1, ..., b_n} \Bigg[ \frac{1}{n(n - 1)} \sum_{i \neq j} \g{ARTS}(b_i, b_j, \rho) \Bigg] \\
    &\quad= \frac{1}{n(n - 1)} \sum_{i \neq j} \E \big[ \g{ARTS}(b_i, b_j, \rho) \big] \\
    &\quad= \frac{1}{n(n - 1)} \sum_{i \neq j} \nabla_\phi \mathcal{E}(\phi) = \nabla_\phi \mathcal{E}(\phi).
\end{align*}
However the computational effort required is $\mathcal{O}(n^2)$, as opposed to LOORF, which is $\mathcal{O}(n)$. But if we restrict ourselves to a symmetric correlation structure, $i.e.$, $\rho_{ij} = \rho$ for $i \neq j$, we can compute the above average in $\mathcal{O}(n)$. To show this, we will make use of the following, proved in Appendix~\ref{appendix:proofs}:
\begin{align}
    \label{eq:podisloorf}
    &\g{LOORF}(\bm b) = \frac{1}{n} \sum_{i=1}^n \Bigg( f(b_i) - \frac{1}{n} \sum_{j=1}^n f(b_j) \Bigg) \nabla_{\phi} \ln p(b_i) \nonumber \\
    &=\frac{1}{n(n - 1)} \resizebox{.76\columnwidth}{!}{$ \displaystyle\sum_{i \neq j} \frac{1}{2} \Big( f(b_i) - f(b_j) \Big) \Big( \nabla_{\phi} \ln p(b_i) - \nabla_{\phi} \ln p(b_j) \Big) $} \nonumber \\
    &= \frac{1}{n(n - 1)} \sum_{i \neq j} \g{PoD}(b_i, b_j),
\end{align}
as well as the fact that $\g{ARTS}(b_i, b_j, \rho) = \g{PoD}(b_i, b_j)/(1 - \rho)$, to obtain the ARMS estimator:
\begin{align*}
    \g{ARMS}(\bm b, \phi, \rho) &= \frac{1}{n(n - 1)} \sum_{i \neq j} \g{ARTS}(b_i, b_j, \rho) \\
    &= \frac{1}{n(n - 1)} \sum_{i \neq j} \frac{\g{PoD}(b_i, b_j)}{1 - \rho} = \frac{\g{LOORF}(\bm b)}{1 - \rho}
\end{align*}
where $\bm b = (b_1, ..., b_n)$. We summarize the derivation in Theorem~\ref{thm:main}. It is clear from Eq.~\ref{eq:podisloorf} why LOORF outperforms (Dis)ARM when the number of samples increases. It is because computing LOORF for $n$ samples is equivalent to averaging $n(n - 1) / 2$ independent pairs, whereas (Dis)ARM uses only $n / 2$ antithetic pairs, and the antithetic variance reduction does not make up for using $n - 1$ times fewer pairs.

\begin{theorem}
    \label{thm:main}
    Let $\tilde{\bm b} = (\tilde b_1, ..., \tilde b_n) \sim \mathcal{B}_n(\sigma(\phi))$ be a sample from an $n$-variate Bernoulli distribution with marginal distributions $\tilde{b}_1, ..., \tilde{b}_n \sim \text{\emph{Bern}}(\sigma(\phi))$ and pairwise correlation $\rho = \text{Corr}(\tilde{b}_i, \tilde{b}_j), i \neq j$. An unbiased estimator of $\nabla_{\phi} \mathbb{E}_{b \sim \text{\emph{Bern}}(\sigma(\phi))} \left[ f(b) \right] $ is:
    \begin{align}
        \label{eq:univariate}
        &\g{\emph{ARMS}}(\tilde{\bm b}, \phi, \rho) \nonumber \\
        &\quad= \frac{1}{n-1} \sum_{i=1}^n \left( f(\tilde{b}_i) - \frac{1}{n} \sum_{j=1}^n f(\tilde{b}_j) \right) \frac{\tilde{b}_i - \sigma(\phi)}{1 - \rho}.
    \end{align}
\end{theorem}

It is simple to show that ARMS generalizes both LOORF and DisARM. For $n$ independent samples $\rho=0$, so the debiasing term becomes $1 / (1 - \rho) = 1$ and the estimator reduces to LOORF. For $n=2$ samples, it reduces to DisARM/U2G if we use an antithetic uniform pair as shown in Theorem~\ref{thm:optimality}.

Although Theorem~\ref{thm:main} shows us how to form an unbiased estimator, it does so assuming we can sample $n$ correlated Bernoulli variables, and that we know their common correlation $\rho$. Therefore, to use ARMS in practice, we must find a way to sample $(b_1, ..., b_n) \sim \mathcal{B}_n(p)$, and also be able to calculate $\rho = \text{Corr}(b_i, b_j)$. The next section outlines two copula based approaches that satisfy both conditions, although there are other ways, whose exploration we leave for future work.

\subsection{Copula Sampling for Multivariate Bernoulli}

A copula sample can be transformed into a multivariate Bernoulli using the reparameterization $b \sim \text{Bern}(p) \iff b = \mathbbm{1}_{u < p}$, $u \sim \text{Unif}(0, 1)$. Furthermore, symmetry in the bivariate copula CDFs:
\begin{align*}
P(u_i < p, u_j < p) &= P(u_k < p, u_l < p), & \forall i \neq j, k \neq l,
\end{align*}
implies symmetry for the Bernoulli correlations $\rho_{ij} = \rho_{kl}$, because $\E[b_i b_j] = P(b_i = 1, b_j = 1) = P(u_i < p, u_j < p)$. As a consequence, if we use a copula for sampling, evaluating the bivariate CDF will be required to calculate~$\rho$ and different copulas will produce different correlations for each $p \in [0, 1]$. Ideally, we want this correlation to be as low as possible for all $p$. For any distribution, the lower limit for a common correlation between $n$ identically distributed variables is $\rho = - 1 / (n - 1)$, which follows from rearranging the non-negativity of the variance equation: $\Var(\sum_{i=1}^n b_i) = n \Var(b_1) + n(n - 1)\rho \Var(b_1) \geq 0.$ We propose two different copulas, both of which start with maximally negatively correlated variables, and preserve most of the correlation when transformed to uniform variables.

\begin{algorithm}[t]
  \caption{Antithetic Dirichlet copula sampling}
  \label{alg:dirichlet_copula}
\begin{algorithmic}
  \STATE {\bfseries Input:} Copula dimension $n$
  \STATE Sample $v_i \sim \text{Unif}(0, 1)$, $i = 1 \dots n$ 
  \STATE Set: $ d_i = \ln(v_i) / \sum_{j=1}^n \ln(v_j)$
  \STATE Invert back to uniform: $\tilde{u}_i = 1 - (1 - d_i)^{n - 1}$
  \STATE {\bfseries return:} $(\tilde u_1, ..., \tilde u_n)$, \; $(1 - \tilde u_1, ..., 1 - \tilde u_n)$
\end{algorithmic}
\end{algorithm}

\subsubsection{Dirichlet Copula}
The first of two sampling approaches for ARMS is a Dirichlet copula, since a sample from the Dirichlet distribution exhibits perfect negative dependence, which means the correlation between any pair of elements in the vector is the lower bound $1 / (n - 1)$. This makes it a suitable candidate, provided we can transform the marginal distributions to uniform, and can compute the bivariate CDF. Both conditions are possible if we restrict ourselves to the uniform Dirichlet density $\bm d \sim \text{Dir}(\bm 1_n)$. In this case, the marginal CDFs of $\bm d$ and $\bm 1 - \bm d$ have closed form solutions $F_{d_i}(x) = 1 - (1 - x)^{n - 1}$ and $F_{1 - d_i}(x) = (1 - x)^{n - 1}$, respectively~\cite{ng2011dirichlet}. Applying Eq.~\ref{eq:cdftransform} results in two different copula samples:
\begin{align*}
    \tilde{\bm u} &= (\tilde{u}_1, ..., \tilde{u}_n), \quad \tilde{u}_i = 1 - (1 - d_i)^{n - 1} \\
    \tilde{\bm u}' &= (\tilde{u}_1', ..., \tilde{u}_n'), \quad \tilde{u}_i' = (1 - d_i)^{n - 1} = 1 - \tilde{u}_i,
\end{align*}
with the full procedure summarized in Algorithm~\ref{alg:dirichlet_copula}. This satisfies the sampling requirement, since given $\tilde{\bm u}$, a multivariate Bernoulli sample is $\tilde{\bm b} = (\tilde{b}_1, ..., \tilde{b}_n)$, where $\tilde{b}_i = \mathbbm{1}_{\tilde u_i < p}$. The bivariate CDF, necessary for calculating $\rho = \text{Corr}(\tilde b_i, \tilde b_j)$, also has a closed form solution:
\begin{align}
    \label{eq:bivdir}
    P(\tilde{u}_i < p, \tilde{u}_j < p) &= P(d_i < q, d_j < q) \nonumber \\ 
    = 2p - 1 + &\max(0, 2(1 - p)^{1 / (n-1)} - 1)^{n - 1} \nonumber \\
    P(\tilde{u}_i' < p, \tilde{u}_j' < p) &= \max(0, 2p^{1 / (n-1)} - 1)^{n - 1},
\end{align}
where $q = 1 - (1 - p)^{1 / (n - 1)}$. We summarize the Dirichlet copula approach with the following theorem, with the full derivation given in Appendix~\ref{appendix:dirichlet}.
\begin{theorem}
    \label{thm:dirichlet_bernoulli}
    Let $\tilde{\bm u}, \tilde{\bm u}'$ be the samples produced by the Dirichlet copula in Algorithm~\ref{alg:dirichlet_copula}. If:
    \begin{align*}
        \tilde{\bm b} &= (\tilde{b}_1, ..., \tilde{b}_n), \quad \tilde{b}_i = \mathbbm{1}_{\tilde u_i < p} \\
        \tilde{\bm b}' &= (\tilde{b}_1', ..., \tilde{b}_n'), \quad \tilde{b}_i' = \mathbbm{1}_{\tilde u'_i < p},
    \end{align*}
    then $\tilde{\bm b}, \tilde{\bm b}' \sim \mathcal{B}_n(p)$ are samples from an $n$-variate Bernoulli distribution with common pairwise correlations:
    \begin{align*}
        \rho &= \frac{\max(0, 2(1 - p)^{\frac{1}{n-1}} - 1)^{n - 1} - (1 - p)^2}{p(1 - p)} \\
        \rho' &= \frac{\max(0, 2p^{1 / (n-1)} - 1)^{n - 1} - p^2 }{p(1 - p)}.
    \end{align*}
\end{theorem}

\begin{figure}[t]
    \vskip 0.2in
    \begin{center}
        \centerline{\includegraphics[width=0.98\columnwidth]{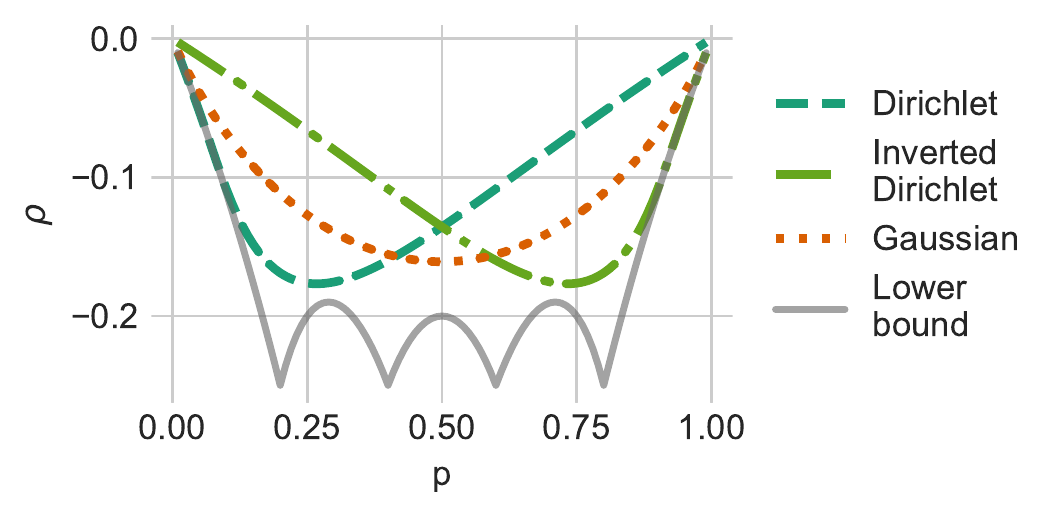}}
        \caption{Correlations $\rho = \text{Corr}(b_i, b_j)$ for $p \in [0, 1]$, when using different copulas to sample a five dimensional multivariate Bernoulli $\bm b \sim \mathcal{B}_5(p)$, along with the bound in~\cite{hoernig2018minimum}.}
        \label{fig:correlations}
    \end{center}
    \vspace{-5mm}
\end{figure}

Lastly, although $\text{Corr}(\tilde{u}_i, \tilde{u}_j) = \text{Corr}(\tilde{u}'_i, \tilde{u}'_j)$, their bivariate CDFs are different, and we observed that if $p > 0.5$, then $\text{Corr}(\tilde{b}_i, \tilde{b}_j) < \text{Corr}(\tilde{b}'_i, \tilde{b}'_j)$ (and vice versa). Therefore, in our experiments we use $\tilde{\bm u}$ when $p > 0.5$ and $\tilde{\bm u}'$ when $p < 0.5$. We illustrate the different Bernoulli correlations of each copula as a function of $p$ in Fig.~\ref{fig:correlations}, along with a lower bound of the lower bound for symmetric Bernoulli variables~\cite{hoernig2018minimum}.

\subsubsection{Antithetic Gaussian Copula}

\begin{algorithm}[!t]
  \caption{Antithetic Gaussian copula sampling}
  \label{alg:gaussian}
\begin{algorithmic}
  \STATE {\bfseries Input:} copula dimension $n$
  \STATE For $i, j = 1, ..., n$, set: 
  $$ (\bm \rho)_{ij} = \begin{cases}  
    -1/(n - 1)&,  i \neq j \\
    1 &, \text{otherwise} 
  \end{cases} $$
  \STATE Sample $\bm x \sim \mathcal{N}(0, \bm \rho)$ 
  \STATE Invert back to uniform: $\tilde{u}_i = \Phi(x_i) \quad i = 1 \dots n$
  \STATE {\bfseries return:} $\tilde u_1, ..., \tilde u_n$
\end{algorithmic}
\end{algorithm}

An alternative sampling method for $n$ dependent uniform variables is the widely used Gaussian copula, Let $\bm x \sim \mathcal{N}(0, \bm \rho)$ denote a sample from an $n$-variate Gaussian, and denote $\Phi(x)$ to be the univariate Gaussian CDF. From Eq~\ref{eq:cdftransform}, by applying the CDF to each dimension of $\bm x$, the result is a Gaussian copula sample: $\bm u = (\Phi(x_1), ... \Phi(x_n))$. Since we are interested in mutually antithetic samples, we use the correlation matrix $\rho_{ij} = -1 / (n - 1), i \neq j$. We summarize the sampling procedure in Algorithm~\ref{alg:gaussian}. The antithetic Gaussian copula can also be recovered from the Dirichlet copula, where $\forall i, \alpha_i = \alpha$, and $\alpha \to \infty$, in which case the distribution becomes a multivariate Gaussian with the same common negative correlation. Although neither the univariate nor bivariate CDF have a closed form, all commonly used deep learning packages contain numerical approximations for both, which we make use of in our experiments.

\subsection{ARMS for the Multivariate Case}
Let $\bm b \sim p(\bm b \,|\, \bm \phi) = \prod_{d=1}^m p_{\phi^d}(b^d)$ denote a $m$-dimensional factorized Bernoulli sample with probabilities $\sigma(\bm \phi) = (\sigma(\bm \phi_1), ..., \sigma(\bm \phi_m))$, such that $b^d \sim \text{Bern}(\sigma(\bm \phi^d)) = p_{\bm \phi^d}(b)$, and let $\bm b^{-d} = (b_1, ..., b_{d-1}, b_{d+1}, ..., b_m)$. We sample $n$ correlated Bernoulli variables for each dimension:
\begin{equation*}
    \tilde{\bm b}^d = (\tilde b_1^d, ..., \tilde b_n^d) \sim \mathcal{B}_n(\sigma(\bm \phi^d)), \quad d = 1, ..., m,
\end{equation*}
with $ \bm \rho^d = \text{Corr}(\tilde{\bm b}^d_i, \tilde{\bm b}^d_j), i \neq j$ being the pairwise correlations. Focusing on the $d^{\text{th}}$ dimension, we have:
\begin{align*}
    &\nabla_{\bm \phi^d} \E_{p_{\bm \phi}(\bm b)} \left[ f(\bm b) \right] = \E_{\bm b^{-d}} \left[ \nabla_{\bm \phi^d} \E_{b^d} \left[ f(\bm b^{-d}, b^d) \right] \right] \\
    &\begin{aligned}
        =\E_{\bm b^{-d}} \Bigg[ \E_{\tilde{\bm b}^d} \Bigg[ &\frac{1}{n-1} \sum_{i=1}^n \Bigg( f(\bm b^{-d}, \tilde{b}_i^d) \\
        &- \frac{1}{n}\sum_{j=1}^n  f(\bm b^{-d}, \tilde{b}_j^d)
        \Bigg) \frac{\tilde{b}_i^d - \sigma(\bm \phi^d)}{1 - \bm \rho^d } \Bigg] \Bigg] .
    \end{aligned}
\end{align*}  
Replacing $\bm b^{-d}$ by the already sampled $\tilde{\bm b}_i^{-d}$, we have:
\begin{align*}
    &\begin{aligned}
        \E_{\tilde{\bm b}^1, ..., \tilde{\bm b}^m} \Bigg[ &\frac{1}{n-1} \sum_{i=1}^n \Bigg( f(\tilde{\bm b}^{-d}_i, \tilde{b}_i^d) \\
        &- \frac{1}{n}\sum_{j=1}^n  f(\tilde{\bm b}^{-d}_j, \tilde{b}_j^d) 
        \Bigg) \frac{\tilde{b}_i^d - \sigma(\bm \phi^d)}{1 - \bm \rho^d } \Bigg]
    \end{aligned} \\
    &\begin{aligned}
        =\E_{\tilde{\bm b}^1, ..., \tilde{\bm b}^m} \Bigg[ &\frac{1}{n-1} \sum_{i=1}^n \Bigg( f(\tilde{\bm b}_i) \\
        &- \frac{1}{n}\sum_{j=1}^n  f(\tilde{\bm b}_j) 
        \Bigg) \frac{\tilde{b}_i^d - \sigma(\bm \phi^d)}{1 - \bm \rho^d } \Bigg],
    \end{aligned}
\end{align*}
where $\tilde{\bm b}_i = (\tilde{\bm b}_i^1, ..., \tilde{\bm b}_i^m)$, i.e. the $i^{\text{th}}$ sample from each dimension. The multivariate ARMS estimator is then:
\begin{align*}
     &\g{ARMS}(\tilde{\bm b}_1, ..., \tilde{\bm b}_n, \bm \phi, \bm \rho) \nonumber \\
     &\quad= \frac{1}{n - 1} \sum_{i=1}^n \Bigg( f(\tilde{\bm b}_i) - \frac{1}{n}\sum_{j=1}^n  f(\tilde{\bm b}_j) \Bigg) \frac{\tilde{\bm b}_i - \sigma(\bm \phi)}{1 - \bm \rho},
\end{align*}
where just like the univariate case, we only need $n$ evaluations of $f$ regardless of the number of dimensions $m$. 

\subsection{ARMS for the Multi Sample Variational Bound}

In variational inference, a commonly optimized objective of the form of Eq.~\ref{eq:factorized} is the ELBO~\cite{jordan1998introduction}, a tractable lower bound of the marginal likelihood:
\begin{equation*}
    \mathcal{L}_{\text{ELBO}} = \E_{q_{\bm \phi}(\bm b | \bm x)} [ \ln p_{\bm \theta}(\bm b, \bm x) - \ln q_{\bm \phi}(\bm b | \bm x) ] \leq  \ln p(\bm x)
.\end{equation*}
If we use multiple independent samples $\bm b_1, ..., \bm b_n$, with $q_{\bm \phi}(\bm b | \bm x) = \prod_{k=1}^n q_{\bm \phi}(\bm b_k | \bm x) $ \cite{burda2016importance} show that we can form a tighter bound:
\begin{equation*}
    \mathcal{L}_n = \E_{q_{\bm \phi}(\bm b | \bm x)} \Bigg[ \ln \Bigg( \frac{1}{n} \sum_{k=1}^n r(\bm b_k) \Bigg) \Bigg], \;r(\bm b_k) = \frac{p_{\bm \theta}(\bm b_k, \bm x)}{q_{\bm \phi}(\bm b_k | \bm x)}
,\end{equation*}
and proved that $\mathcal{L}_{\text{ELBO}} \leq \mathcal{L}_{n} \leq \mathcal{L}_{n+1} \leq \ln p(\bm x)$. The multi sample bound depends non-linearly on each sample due to the logarithm inside the expectation, which means replacing the $n$ i.i.d. samples with dependent ones changes the objective. For example, if we let $\bm b_1 = ... = \bm b_n = \bm b$, then instead of $\mathcal{L}_n$, we obtain:
\begin{equation*}
    \resizebox{.96\columnwidth}{!}{$ \displaystyle  \E_{\bm b \sim q_{\bm \phi}(\bm b | \bm x)} \left[ \ln  \frac{1}{n} \sum_{i=1}^n r(\bm b)  \right] = \E_{\bm b \sim q_{\bm \phi}(\bm b | \bm x)} \Big[ \ln r(\bm b) \Big] = \mathcal{L}_1$}.
\end{equation*}
However, we can still use ARMS to create a local baseline for each sample if we draw $n$ additional correlated samples. Let $\bm b_1, .., \bm b_n$ denote the i.i.d. samples and $\tilde{\bm b}_1, .., \tilde{\bm b}_n$ the correlated samples. Define $f(\bm b) = \ln \left( \frac{1}{n} \sum_{k=1}^n r(\bm b_k) \right)$, and $f_{\bm b_{-k}}(\bm b_k) = \ln \left( \frac{1}{n} \big( \sum_{l \neq k} r(\bm b_l) + r(\bm b_k) \big) \right)$. Because the samples are i.i.d., the gradient can be written as:
\begin{align*}
    &\nabla_{\bm \phi} \mathbb{E}_{\bm b}[ f(\bm b)] = \mathbb{E}_{\bm b} \Big[ f(\bm b) 
    \nabla_{\bm \phi} \ln \prod_{k=1}^n q_{\bm \phi}(\bm b_k | \bm x) \Big] = \\
    &= \sum_{k=1}^n \mathbb{E}_{\bm b} \Big[ f(\bm b) \nabla_{\bm \phi} \ln q_{\bm \phi}(\bm b_k | \bm x) \Big] = \\ 
    &= \sum_{k=1}^n \mathbb{E}_{\bm b_{-k}} \Bigg[ \mathbb{E}_{\bm b_k} \Big[ f_{\bm b_{-k}}(\bm b_k) \nabla_{\bm \phi} \ln q_{\bm \phi}(\bm b_k | \bm x) \Big] \Bigg] \\
    &= \sum_{k=1}^n \mathbb{E}_{\bm b_{-k}} \Bigg[ \mathbb{E}_{\tilde{\bm b}_1, .., \tilde{\bm b}_n} \bigg[ \frac{1}{n-1} \sum_{i=1}^n \Big( f_{\bm b_{-k}}(\tilde{\bm b}_i) \\
    &\hspace{8em} - \frac{1}{n} \sum_{j=1}^n f_{\bm b_{-k}}(\tilde{\bm b}_j) \Big) \frac{\tilde{\bm b}_i - \sigma(\bm \phi)}{1 - \bm \rho} \bigg] \Bigg],
\end{align*}
where in the last line we replaced the inner expectation, which is just REINFORCE w.r.t. $\bm b_k$, with the ARMS estimator. Note that we can precompute $r(\bm b_1), ..., r(\bm b_n)$ and $r(\tilde{\bm b}_1), ..., r(\tilde{\bm b}_n)$ for a total of $2n$ function evaluations, which is the same number required by DisARM for optimizing the same $n$ sample bound, and the number required by VIMCO for the $2n$ sample bound, which is the what we compare to in the experiments.

\begin{figure*}[!th]
    \begin{center}
        \centerline{\includegraphics[width=1.98\columnwidth]{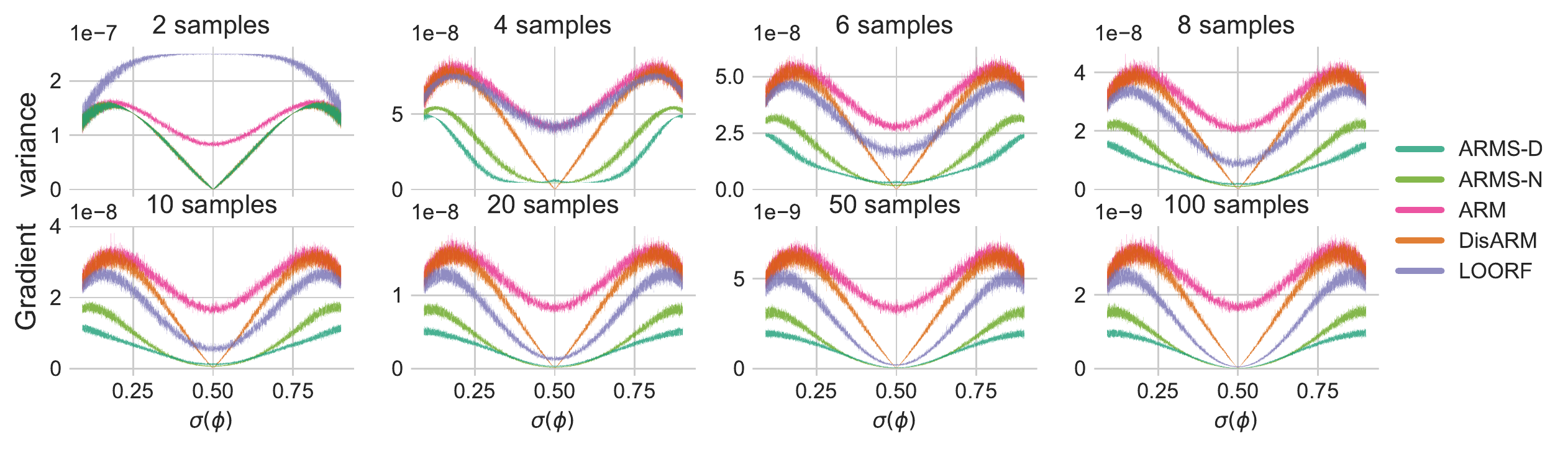}}
        \vspace{-4.5mm}
        \caption{Variance of the gradients of each estimator for the toy problem across a range of values of $\phi$. The top and bottom row correspond to using $n \in \{2, 4, 6, 8\}$ and $n \in \{10, 20, 50, 100\}$ samples, respectively, at each gradient step. The variance is estimated using 1000 Monte Carlo samples per step.}
        \label{fig:toy}
    \end{center}
    \vspace{-5mm}
    \begin{center}
        \centerline{\includegraphics[width=1.9\columnwidth]{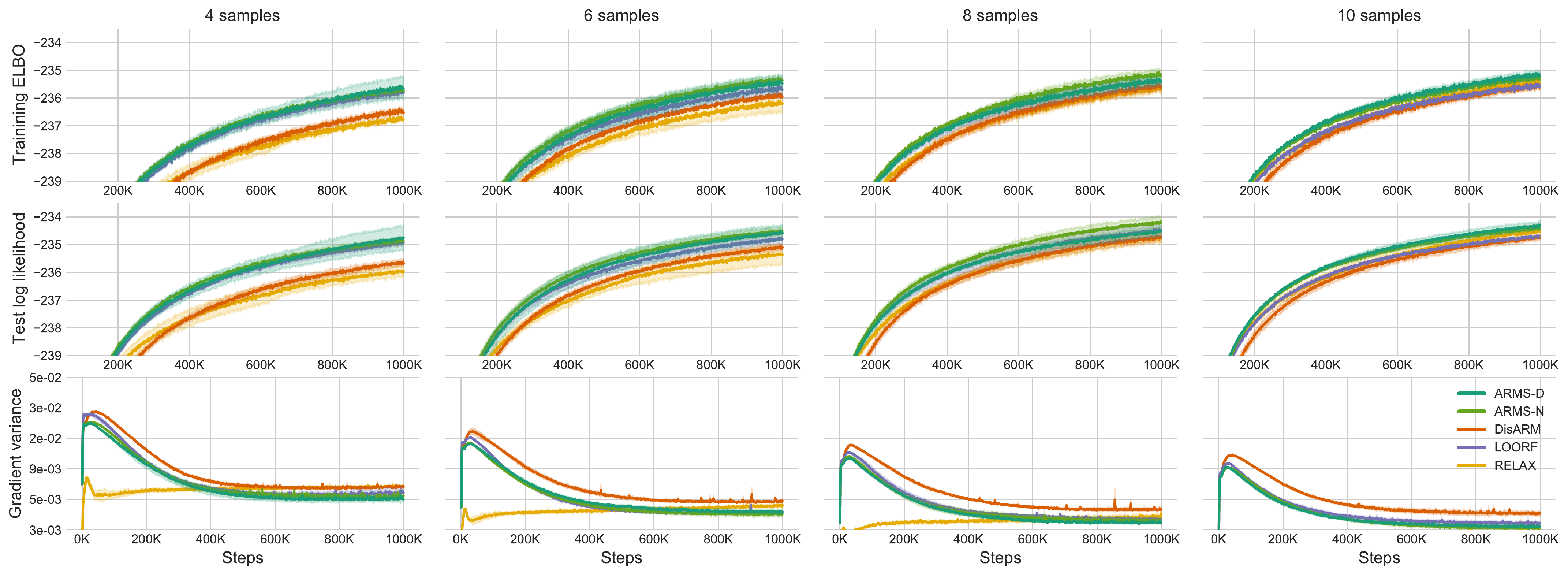}}
       \vspace{-4mm}
        \caption{Training a nonlinear discrete VAE on Fashion MNIST using the ELBO. Columns correspond to $n \in \{4, 6, 8, 10 \}$ samples used per step, respectively. Rows correspond to the training ELBO, test log likelihood, and the variance of the gradient updates averaged over all parameters. Results for different datasets and other networks can be found in Appendix~\ref{appendix:results}.}
        \label{fig:training}
    \end{center}
    \vspace{-7mm}
\end{figure*}

\begin{table*}[!th]
    
    \caption{Final training ELBO of VAEs using different estimators. Results are reported on three datasets: Dynamic MNIST, Fashion MNIST, and Omniglot, and for 4, 6, 8, and 10 samples, with the best performing methods in bold.}
    \label{tab:results}
    \begin{center}
    \begin{small}
    \begin{sc}
    \resizebox{1.75\columnwidth}{!}{
    \begin{tabular}{ccc|rrrrr}
    \toprule
    \multicolumn{3}{c}{Samples} & \multicolumn{1}{c}{ARMS-D} &  \multicolumn{1}{c}{ARMS-N} & \multicolumn{1}{c}{LOORF} & \multicolumn{1}{c}{DisARM} & \multicolumn{1}{c}{RELAX} \\
    \midrule
    \multirow{8}{*}{\rotatebox[origin=c]{90}{\textbf{Dynamic MNIST}}} & \multirow{4}{*}{\rotatebox{90}{Linear}} & 4 &   -112.13 ± 0.10 &  \textbf{-111.96} ± \textbf{0.09} & -112.32 ± 0.04 &  -113.26 ± 0.05 &  -112.98 ± 0.25 \\
    &           & 6 &  -111.03 ± 0.02 &  -\textbf{110.89} ± \textbf{0.07} &  -110.99 ± 0.07 &  -112.11 ± 0.03 &  -111.46 ± 0.06 \\
    &           & 8 &   \textbf{-110.30} ± \textbf{0.04} &  -110.62 ± 0.06 &  -110.42 ± 0.04 &  -111.78 ± 0.07 &  -110.58 ± 0.01 \\
    &           & 10 &  \textbf{-110.08} ± \textbf{0.05} &  -110.14 ± 0.09 &  -110.17 ± 0.04 &  -111.08 ± 0.11 &  -110.17 ± 0.09 \\ \cmidrule{3-8}
    & \multirow{4}{*}{\rotatebox[origin=c]{90}{Nonlinr}} & 4 &   \textbf{-98.65} ± \textbf{0.16} &   -98.97 ± 0.13 &   \textbf{-98.62} ± \textbf{0.05} &  -100.45 ± 0.16 &  -100.52 ± 0.08 \\
    &           & 6 &   -98.53 ± 0.13 &   \textbf{-97.87} ± \textbf{0.01} &   -98.14 ± 0.18 &   -99.28 ± 0.11 &   -99.17 ± 0.17 \\
    &           & 8 & \textbf{-97.90} ± \textbf{0.12} &    \textbf{-97.89} ± \textbf{0.10} &   -98.14 ± 0.21 &   -98.69 ± 0.21 &    -98.80 ± 0.02 \\
    &           & 10 &   -97.64 ± 0.06 &   \textbf{-97.32} ± \textbf{0.11} &    -97.50 ± 0.29 &   -98.62 ± 0.12 &   -98.69 ± 0.07 \\
    \midrule
    \end{tabular}}
    \resizebox{1.75\columnwidth}{!}{
    \begin{tabular}{ccc|rrrrr}
        \multirow{8}{*}{\rotatebox[origin=c]{90}{\textbf{Fashion MNIST}}} & \multirow{4}{*}{\rotatebox[origin=c]{90}{Linear}} & 4 &  \textbf{-252.56} ± \textbf{0.11} &  -252.69 ± 0.06 &   -252.71 ± 0.09 &  -254.02 ± 0.05 &  -253.53 ± 0.06 \\
        &           & 6 &  -251.94 ± 0.13 &  \textbf{-251.73} ± \textbf{0.05} &  -252.03 ± 0.08 &  -252.97 ± 0.06 &  -252.31 ± 0.14 \\
        &           & 8 &  \textbf{-251.32} ± \textbf{0.11} &  \textbf{-251.11} ± \textbf{0.23} &   -251.41 ± 0.10 &  -252.57 ± 0.05 &  -251.36 ± 0.08 \\
        &           & 10 &  -251.29 ± 0.02 &  \textbf{-251.08} ± \textbf{0.08} &   -251.26 ± 0.03 &  -251.75 ± 0.21 &  \textbf{-251.16} ± \textbf{0.06} \\ \cmidrule{3-8}
        & \multirow{4}{*}{\rotatebox[origin=c]{90}{Nonlinr}} & 4 &   \textbf{-235.65} ± \textbf{0.12} &  -235.75 ± 0.06 &   -235.80 ± 0.07 &  -236.54 ± 0.06 &  -236.77 ± 0.03 \\
        &           & 6 &  -235.47 ± 0.19 &  \textbf{-235.36} ± \textbf{0.08} &   -235.70 ± 0.13 &  -235.94 ± 0.05 &   -236.20 ± 0.25 \\
        &           & 8 &   -235.41 ± 0.10 &  \textbf{-235.19} ± \textbf{0.14} &   -235.40 ± 0.13 &  -235.62 ± 0.16 &   -235.70 ± 0.14 \\
        &           & 10 &  \textbf{-235.18} ± \textbf{0.11} &  -235.32 ± 0.05 &  -235.59 ± 0.01 &   -235.60 ± 0.09 &  -235.46 ± 0.19 \\
    \midrule
    \end{tabular}}
    \resizebox{1.75\columnwidth}{!}{
    \begin{tabular}{ccc|rrrrr}
        \multirow{8}{*}{\rotatebox[origin=c]{90}{\textbf{Omniglot}}} & \multirow{4}{*}{\rotatebox[origin=c]{90}{Linear}} & 4 &  \textbf{-118.25} ± \textbf{0.08} &  \textbf{-118.27} ± \textbf{0.05} &  -118.41 ± 0.07 &  -119.24 ± 0.17 &  -118.75 ± 0.08 \\
          &           & 6 &  \textbf{-117.62} ± \textbf{0.01} &  \textbf{-117.62} ± \textbf{0.04} &  -117.75 ± 0.08 &  -118.47 ± 0.12 &   -117.90 ± 0.03 \\
          &           & 8 &   \textbf{-117.60} ± \textbf{0.05} &  -117.66 ± 0.12 &   -117.74 ± 0.10 &   -118.41 ± 0.10 &  -117.71 ± 0.02\\
          &           & 10 &  \textbf{-117.03} ± \textbf{0.09} &  \textbf{-116.99} ± \textbf{0.04} &   -117.21 ± 0.08 &   -117.70 ± 0.01 &  -117.13 ± 0.05 \\ \cmidrule{3-8}  
        & \multirow{4}{*}{\rotatebox[origin=c]{90}{Nonlinr}} & 4 &  -112.09 ± 0.27 &  \textbf{-112.03} ± \textbf{0.12} &   -112.20 ± 0.26 &  -113.24 ± 0.16 &  -114.08 ± 0.35 \\
        &           & 6 &   -111.50 ± 0.06 &   \textbf{-111.39} ± \textbf{0.10} &  \textbf{-111.26} ± \textbf{0.15} &   -112.30 ± 0.05 &  -113.71 ± 0.13 \\
        &           & 8 &  \textbf{-110.91} ± \textbf{0.04} &   -111.01 ± 0.06 &  \textbf{-110.85} ± \textbf{0.35} &  -111.82 ± 0.09 &   -113.64 ± 0.10 \\
        &           & 10 &  \textbf{-110.66} ± \textbf{0.05} &  -110.79 ± 0.26 &   -110.79 ± 0.20 &  -111.33 ± 0.19 &    -114.00 ± 0.10 \\
    \bottomrule
    \end{tabular}}
    \end{sc}
    \end{small}
    \end{center}
    \vspace{-4mm}
\end{table*}

\section{Experimental Results}

\begin{figure*}[!th]
    \begin{center}
        \centerline{\includegraphics[width=1.9\columnwidth]{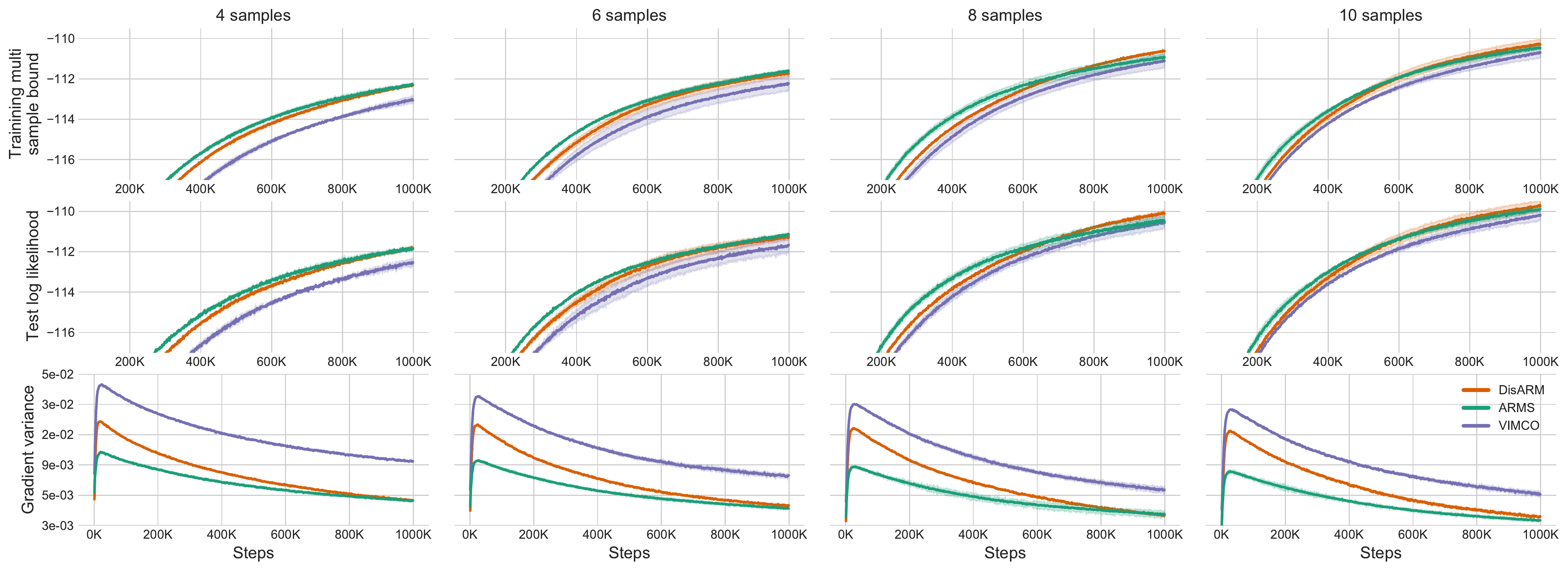}}
       \vspace{-4.5mm}
        \caption{Training a linear discrete VAE on Dynamic MNIST using the multi sample bound. Columns correspond to $n \in \{4, 6, 8, 10 \}$ samples used per step, respectively. Rows correspond to the training multi sample bound, test log likelihood, and the variance of the gradient updates averaged over all parameters. Results for different datasets and other networks can be found in Appendix~\ref{appendix:results}.}
        \label{fig:mstraining}
    \end{center}
    \vspace{-4.5mm}
\end{figure*}
First, to illustrate the benefit of jointly antithetic samples, we compare ARMS, using either a Dirichlet (ARMS-D) or Gaussian copula (ARMS-N), to ARM, DisARM, and LOORF on the following toy problem~\cite{tucker2017rebar}, where the task is to maximize:
\begin{equation*}
    \mathcal{E}(\phi) = \E_{b} [ (b - 0.499)^2 ], \quad b \sim \text{Bern}(\sigma(\phi))
,\end{equation*}
with the optimal solution being $\sigma(\phi) = 1$. We run the analysis for $n \in \{2, 4, 6, 8, 10, 20, 50, 100\}$ samples per gradient step, using an even number to ensure a fair comparison to methods that can only use pairs.  Figure~\ref{fig:toy} shows the variance of the gradient of each estimator as the optimization progresses from $\sigma(\phi)=0.1$ to $\sigma(\phi)=0.9$, which takes approximately 20,000 steps. Both ARMS-D and ARMS-N have significantly lower variance for almost all values of $\phi$, but the Dirichlet copula appears to have lower variance, except for a narrow range near $\sigma(\phi)=0.5$. Interestingly, the variance reduction for both is still large even for $n=10$ samples, when the correlation between each Bernoulli pair in the sample cannot be larger than $-1/9$.

\subsection{ELBO Based Discrete VAE}
\label{sec:elbo}
Our experimental setup follows the one in \citet{yin2019arm} and \citet{dong2020disarm}, and all VAE experiments are built on top of the available DisARM code. For this task, we compare ARMS-D and ARMS-N to DisARM/U2G, LOORF, and RELAX. We omit ARM, since it is outperformed by DisARM/U2G, as shown in both \citet{yin2020probabilistic} and \citet{dong2020disarm}. All estimators optimize the ELBO of a variational autoencoder (VAE)~\cite{kingma2013auto}, a commonly used method for generative modelling. The comparison is done on three different benchmark datasets: dynamically binarized MNIST, Fashion MNIST, and Omniglot, with each dataset split into the training, validation, and test sets. For each dataset we use either a linear or nonlinear encoder decoder pairs, with $n \in \{4, 6, 8, 10 \}$ samples per gradient step. When $n=2$, both copulas reduce to DisARM/U2G, for which our experiments mirrored the results found in~\citet{dong2020disarm}, so we omit this case. All results are reported based on five independent runs.

The implementation details are the following: each VAE contains a stochastic binary layer with 200 units, with two types of encoder-decoder pairs used: linear or nonlinear. The nonlinear network has two hidden layers of 200 units each, using LeakyReLU~\cite{maas2013rectifier} activations with a coefficient of 0.3. Adam~\cite{kingma2015adam} with a learning rate of $1e^{-4}$ was used to optimize the network parameters, and SGD with learning rate $1e^{-2}$ for the prior distribution logits. The optimization is run for $10^6$ steps with mini batches of size 50. For RELAX, the scaling factor is initialized to 1, the temperature to 0.1, and the control variate is a neural network with one hidden layer of 137 units using LeakyReLU activations. The only data preprocessing includes subtracting the global mean of the dataset from each image. All the models were trained on a K40 Nvidia GPU and Intel Xeon E5-2680 processor.

In Fig.~\ref{fig:training}, we plot the train ELBO, test log likelihood, and variance, for a nonlinear network trained on Fashion MNIST. The variance at each step is an average of the gradient variance of all networks parameters. The same visualization for the other datasets and linear networks can be found in Appendix~\ref{appendix:results}. In Table~\ref{tab:results}, we report the average of five final training ELBO results for all estimators on the three datasets, both types of networks, and different number of samples. The corresponding test log likelihood table can be found in Appendix~\ref{appendix:results}, with the performance gaps between methods being similar. ARMS consistently outperforms the state of the art and using either the Dirichlet and Gaussian copula appears to have no discernible difference. For linear networks, RELAX is competitive, thought not for nonlinear networks. LOORF generally appears to be a strong overlooked estimator that closely tails the performance of ARMS, indicating that it is beneficial to use all samples when creating a baseline, unlike pair based methods.

\begin{table}[!t]
    \caption{Final train multi sample bound of a Discrete VAE trained with different estimators. Results are averaged over five runs, with the best performing methods in bold.}
    \label{tab:msresults}
    \vskip -0.25in
    \begin{center}
    \begin{small}
    \begin{sc}
    \setlength\tabcolsep{3.5pt}
    \resizebox{0.9\columnwidth}{!}{
    \begin{tabular}{ccc|rrr}
    \toprule
    \multicolumn{3}{c}{Samples} & \multicolumn{1}{c}{ARMS} &  \multicolumn{1}{c}{DisARM} & \multicolumn{1}{c}{VIMCO} \\
    \midrule
         \multirow{8}{*}{\rotatebox[origin=c]{90}{\textbf{Dynamic MNIST}}} & \multirow{4}{*}{\rotatebox{90}{Linear}} & 4 &  \textbf{-112.31} ± \textbf{0.05} &  -113.26 ± 0.05 &  -113.07 ± 0.14 \\
         &           & 6 &  \textbf{-111.73} ± \textbf{0.04} &  -112.11 ± 0.03 &  -112.23 ± 0.27 \\
         &           & 8 &  \textbf{-110.62} ± \textbf{0.13} &  -111.78 ± 0.07 &  -111.10 ± 0.27 \\
         &           & 10 &   \textbf{-110.30} ± \textbf{0.07} &  -111.08 ± 0.11 &   -110.69 ± 0.20 \\ \cmidrule{3-6}  
         & \multirow{4}{*}{\rotatebox[origin=c]{90}{Nonlinr}} & 4 &    \textbf{-98.00} ± \textbf{0.02} &  -100.45 ± 0.16 &   -98.07 ± 0.06 \\
         &           & 6 &   \textbf{-96.63} ± \textbf{0.15} &   -99.28 ± 0.11 &   -96.91 ± 0.12 \\
         &           & 8 &   \textbf{-96.17} ± \textbf{0.09} &   -98.69 ± 0.21 &   -96.49 ± 0.04 \\
         &           & 10 &    \textbf{-95.27} ± \textbf{0.10} &   -98.62 ± 0.12 &   -95.77 ± 0.03 \\
    \midrule
    \end{tabular}
    } \resizebox{0.9\columnwidth}{!}{
    \begin{tabular}{ccc|rrr}
         \multirow{8}{*}{\rotatebox[origin=c]{90}{\textbf{Fashion MNIST}}} & \multirow{4}{*}{\rotatebox[origin=c]{90}{Linear}} & 4 &  \textbf{-252.24} ± \textbf{0.18} &  -254.02 ± 0.05 &  -252.75 ± 0.06 \\
         &           & 6 &   \textbf{-251.40} ± \textbf{0.01} &  -252.97 ± 0.06 &  -251.64 ± 0.19 \\
         &           & 8 &  \textbf{-250.68} ± \textbf{0.15} &  -252.57 ± 0.05 &  -250.78 ± 0.04 \\
         &           & 10 &  \textbf{-250.54} ± \textbf{0.04} &  -251.75 ± 0.21 &  -250.67 ± 0.12 \\ \cmidrule{3-6}  
         & \multirow{4}{*}{\rotatebox[origin=c]{90}{Nonlinr}} & 4 &   \textbf{-234.69} ± \textbf{0.10} &  -236.54 ± 0.06 &   -234.71 ± 0.20 \\
         &           & 6 &  \textbf{-233.68} ± \textbf{0.15} &  -235.94 ± 0.05 &  \textbf{-233.72} ± \textbf{0.45} \\
         &           & 8 &  \textbf{-233.05} ± \textbf{0.02} &  -235.62 ± 0.16 &    -233.40 ± 0.08 \\
         &           & 10 &  \textbf{-232.74} ± \textbf{0.05} &   -235.60 ± 0.09 &  -233.21 ± 0.07 \\
    \midrule
    \end{tabular}
    } \resizebox{0.9\columnwidth}{!}{
    \begin{tabular}{ccc|rrr}
      \multirow{8}{*}{\rotatebox[origin=c]{90}{\textbf{Omniglot}}} & \multirow{4}{*}{\rotatebox[origin=c]{90}{Linear}} & 4 &   \textbf{-119.23} ± \textbf{0.04} &  \textbf{-119.24} ± \textbf{0.17} &  -120.16 ± 0.06 \\
         &           & 6 &   \textbf{-118.50} ± \textbf{0.23} &  \textbf{-118.47} ± \textbf{0.12} &  -119.07 ± 0.08 \\
         &           & 8 &   \textbf{-118.40} ± \textbf{0.09} &  -118.02 ± 0.08 &  \textbf{-118.49} ± \textbf{0.11} \\
         &           & 10 &  \textbf{-118.19} ± \textbf{0.03} &  -117.70 ± 0.01 &  -118.35 ± 0.14 \\ \cmidrule{3-6}  
         & \multirow{4}{*}{\rotatebox[origin=c]{90}{Nonlinr}} & 4 &   \textbf{-111.56} ± \textbf{0.10} &  -113.24 ± 0.16 &  -111.72 ± 0.18 \\
         &           & 6 &   \textbf{-110.51} ± \textbf{0.06} &  -112.30 ± 0.05 &  -110.62 ± 0.08 \\
         &           & 8 &   \textbf{-109.76} ± \textbf{0.10} &  -111.82 ± 0.09 &  -109.84 ± 0.01 \\
         &           & 10 &  \textbf{-109.35} ± \textbf{0.01} &  -111.33 ± 0.19 &  -109.53 ± 0.06 \\
    \bottomrule
    \end{tabular}
    }
    \end{sc}
    \end{small}
    \end{center}
    \vspace{-3mm}
\end{table}

\subsection{Multi Sample Bound Based Discrete VAE}
For the multi sample bound objective, we compare the multi sample version of ARMS using a Dirichlet copula to VIMCO~\cite{mnih2016variational}, an estimator tailored to this objective, and the multi sample version of DisARM~\cite{dong2020disarm}. To ensure a fair comparison, all three estimators use the same number of function evaluations, which means VIMCO optimizes the $n$-sample bound, whereas ARMS and DisARM optimize the $\frac{n}{2}$-sample bound. All the experimental details are otherwise  identical to Section~\ref{sec:elbo}. We show training plots over time in Fig.~\ref{fig:mstraining}. To be able to compare the variance, we average two $n$-sample bound VIMCO estimates instead, so that the objective optimized is the same. The average of five final train multi sample bounds are shown in Table~\ref{tab:msresults}, for each dataset, both types of networks and different number of samples. Similarly to the ELBO case, ARMS outperforms both VIMCO and DisARM, regardless of the dataset, the type of network, or the number of samples.

\section{Conclusion}

To optimize the parameters of Bernoulli variables in expectation-based objectives, we proposed ARMS, an unbiased, low-variance gradient estimator based on $n$ jointly antithetic samples. Also presented are two ways of generating $n$ jointly antithetic Bernoulli samples, based on a Dirichlet copula and a Gaussian copula, respectively. For $n=2$ samples, both copulas produce the same antithetic used by DisARM/U2G. ARMS also generalizes LOORF, which can be obtained by using $n$ independent samples. As shown by the experiments, when training a variational autoencoder using either the ELBO or the multi-sample bound, ARMS outperforms both estimators based on multiple independent samples, and the ones based on pairs of antithetic samples. There are several potential avenues of future work. We showcase two different ways of sampling $n$ jointly antithetic Bernoulli variables using either a Dirichlet or Gaussian copula, but there could be better ways, not necessarily based on copulas. Another promising direction is extending ARMS to categorical variables, which has been done for ARM to arrive at ARSM \cite{yin2019arsm}. Lastly, we optimized the multi sample bound with i.i.d. samples for fair comparison, but using antithetic samples is also a lower bound of the marginal likelihood, albeit a different bound.

\section*{Acknowledgements}
We thank the authors of DisARM for publicly releasing their code. We also acknowledge the computational support of the Texas Advanced Computing Center, and we thank the reviewers for their thoughtful feedback.

\bibliography{references}

\begin{thebibliography}{37}
\providecommand{\natexlab}[1]{#1}
\providecommand{\url}[1]{\texttt{#1}}
\expandafter\ifx\csname urlstyle\endcsname\relax
  \providecommand{\doi}[1]{doi: #1}\else
  \providecommand{\doi}{doi: \begingroup \urlstyle{rm}\Url}\fi

\bibitem[Bengio et~al.(2013)Bengio, L{\'e}onard, and
  Courville]{bengio2013estimating}
Bengio, Y., L{\'e}onard, N., and Courville, A.
\newblock Estimating or propagating gradients through stochastic neurons for
  conditional computation.
\newblock \emph{arXiv preprint arXiv:1308.3432}, 2013.

\bibitem[Burda et~al.(2016)Burda, Grosse, and
  Salakhutdinov]{burda2016importance}
Burda, Y., Grosse, R.~B., and Salakhutdinov, R.
\newblock Importance weighted autoencoders.
\newblock In \emph{4th International Conference on Learning Representations,
  {ICLR}}, 2016.

\bibitem[Dong et~al.(2020)Dong, Mnih, and Tucker]{dong2020disarm}
Dong, Z., Mnih, A., and Tucker, G.
\newblock Disarm: An antithetic gradient estimator for binary latent variables.
\newblock In \emph{Advances in Neural Information Processing Systems 33}, 2020.

\bibitem[Fu(2006)]{fu2006gradient}
Fu, M.~C.
\newblock Gradient estimation.
\newblock \emph{Handbooks in operations research and management science},
  13:\penalty0 575--616, 2006.

\bibitem[Glynn(1990)]{glynn1990likelihood}
Glynn, P.~W.
\newblock Likelihood ratio gradient estimation for stochastic systems.
\newblock \emph{Communications of the ACM}, 33\penalty0 (10):\penalty0 75--84,
  1990.

\bibitem[Grathwohl et~al.(2018)Grathwohl, Choi, Wu, Roeder, and
  Duvenaud]{grathwohl2018relax}
Grathwohl, W., Choi, D., Wu, Y., Roeder, G., and Duvenaud, D.
\newblock Backpropagation through the void: Optimizing control variates for
  black-box gradient estimation.
\newblock In \emph{6th International Conference on Learning Representations,
  {ICLR}}, 2018.

\bibitem[Gu et~al.(2016)Gu, Levine, Sutskever, and Mnih]{gu2015muprop}
Gu, S., Levine, S., Sutskever, I., and Mnih, A.
\newblock Muprop: Unbiased backpropagation for stochastic neural networks.
\newblock In Bengio, Y. and LeCun, Y. (eds.), \emph{4th International
  Conference on Learning Representations, {ICLR}}, 2016.

\bibitem[Hoernig(2018)]{hoernig2018minimum}
Hoernig, S.
\newblock On the minimum correlation between symmetrically distributed random
  variables.
\newblock \emph{Operations Research Letters}, 46\penalty0 (4):\penalty0
  469--471, 2018.

\bibitem[Jang et~al.(2017)Jang, Gu, and Poole]{jang2017categorical}
Jang, E., Gu, S., and Poole, B.
\newblock Categorical reparameterization with gumbel-softmax.
\newblock In \emph{5th International Conference on Learning Representations,
  {ICLR}}. OpenReview.net, 2017.

\bibitem[Jordan et~al.(1998)Jordan, Ghahramani, Jaakkola, and
  Saul]{jordan1998introduction}
Jordan, M.~I., Ghahramani, Z., Jaakkola, T.~S., and Saul, L.~K.
\newblock An introduction to variational methods for graphical models.
\newblock In \emph{Learning in graphical models}, pp.\  105--161. Springer,
  1998.

\bibitem[Kingma \& Ba(2015)Kingma and Ba]{kingma2015adam}
Kingma, D.~P. and Ba, J.
\newblock Adam: {A} method for stochastic optimization.
\newblock In Bengio, Y. and LeCun, Y. (eds.), \emph{3rd International
  Conference on Learning Representations, {ICLR}}, 2015.

\bibitem[Kingma \& Welling(2014)Kingma and Welling]{kingma2013auto}
Kingma, D.~P. and Welling, M.
\newblock Auto-encoding variational bayes.
\newblock In Bengio, Y. and LeCun, Y. (eds.), \emph{2nd International
  Conference on Learning Representations, {ICLR}}, 2014.

\bibitem[Kool et~al.(2019)Kool, van Hoof, and Welling]{kool2019buy}
Kool, W., van Hoof, H., and Welling, M.
\newblock Buy 4 {REINFORCE} samples, get a baseline for free!
\newblock In \emph{Workshop, Deep Reinforcement Learning Meets Structured
  Prediction, {ICLR}}, 2019.

\bibitem[Kucukelbir et~al.(2017)Kucukelbir, Tran, Ranganath, Gelman, and
  Blei]{kucukelbir2017automatic}
Kucukelbir, A., Tran, D., Ranganath, R., Gelman, A., and Blei, D.~M.
\newblock Automatic differentiation variational inference.
\newblock \emph{The Journal of Machine Learning Research}, 18\penalty0
  (1):\penalty0 430--474, 2017.

\bibitem[Lorberbom et~al.(2019)Lorberbom, Jaakkola, Gane, and
  Hazan]{lorberbom2018direct}
Lorberbom, G., Jaakkola, T.~S., Gane, A., and Hazan, T.
\newblock Direct optimization through arg max for discrete variational
  auto-encoder.
\newblock In Wallach, H.~M., Larochelle, H., Beygelzimer, A.,
  d'Alch{\'{e}}{-}Buc, F., Fox, E.~B., and Garnett, R. (eds.), \emph{Advances
  in Neural Information Processing Systems 32}, pp.\  6200--6211, 2019.

\bibitem[Maas et~al.(2013)Maas, Hannun, and Ng]{maas2013rectifier}
Maas, A.~L., Hannun, A.~Y., and Ng, A.~Y.
\newblock Rectifier nonlinearities improve neural network acoustic models.
\newblock In \emph{ICML Workshop on Deep Learning for Audio, Speech and
  Language Processing}. Citeseer, 2013.

\bibitem[Maddison et~al.(2017)Maddison, Mnih, and Teh]{maddison2017concrete}
Maddison, C.~J., Mnih, A., and Teh, Y.~W.
\newblock The concrete distribution: {A} continuous relaxation of discrete
  random variables.
\newblock In \emph{5th International Conference on Learning Representations,
  {ICLR}}. OpenReview.net, 2017.

\bibitem[Mnih \& Gregor(2014)Mnih and Gregor]{mnih2014neural}
Mnih, A. and Gregor, K.
\newblock Neural variational inference and learning in belief networks.
\newblock In \emph{International Conference on Machine Learning}, pp.\
  1791--1799. PMLR, 2014.

\bibitem[Mnih \& Rezende(2016)Mnih and Rezende]{mnih2016variational}
Mnih, A. and Rezende, D.~J.
\newblock Variational inference for monte carlo objectives.
\newblock In \emph{Proceedings of the 33nd International Conference on Machine
  Learning, {ICML}}, volume~48, pp.\  2188--2196. JMLR.org, 2016.

\bibitem[Mohamed et~al.(2020)Mohamed, Rosca, Figurnov, and
  Mnih]{mohamed2020monte}
Mohamed, S., Rosca, M., Figurnov, M., and Mnih, A.
\newblock Monte carlo gradient estimation in machine learning.
\newblock \emph{J. Mach. Learn. Res.}, 21:\penalty0 132:1--132:62, 2020.

\bibitem[Ng et~al.(2011)Ng, Tian, and Tang]{ng2011dirichlet}
Ng, K.~W., Tian, G.-L., and Tang, M.-L.
\newblock \emph{Dirichlet and related distributions: Theory, methods and
  applications}, volume 888.
\newblock John Wiley \& Sons, 2011.

\bibitem[Owen(2013)]{owen2013monte}
Owen, A.~B.
\newblock Monte carlo theory, methods and examples.
\newblock \emph{,}, 2013.

\bibitem[Paisley et~al.(2012)Paisley, Blei, and Jordan]{paisley2012variational}
Paisley, J.~W., Blei, D.~M., and Jordan, M.~I.
\newblock Variational bayesian inference with stochastic search.
\newblock In \emph{Proceedings of the 29th International Conference on Machine
  Learning, {ICML}}, 2012.

\bibitem[Ranganath et~al.(2014)Ranganath, Gerrish, and
  Blei]{ranganath2014black}
Ranganath, R., Gerrish, S., and Blei, D.
\newblock Black box variational inference.
\newblock In \emph{Artificial intelligence and statistics}, pp.\  814--822.
  PMLR, 2014.

\bibitem[Ren et~al.(2019)Ren, Zhao, and Ermon]{ren2019adaptive}
Ren, H., Zhao, S., and Ermon, S.
\newblock Adaptive antithetic sampling for variance reduction.
\newblock In \emph{International Conference on Machine Learning}, pp.\
  5420--5428. PMLR, 2019.

\bibitem[Rezende et~al.(2014)Rezende, Mohamed, and
  Wierstra]{rezende2014stochastic}
Rezende, D.~J., Mohamed, S., and Wierstra, D.
\newblock Stochastic backpropagation and approximate inference in deep
  generative models.
\newblock In \emph{International conference on machine learning}, pp.\
  1278--1286. PMLR, 2014.

\bibitem[Rosca et~al.(2019)Rosca, Figurnov, Mohamed, and
  Mnih]{rosca2019measure}
Rosca, M., Figurnov, M., Mohamed, S., and Mnih, A.
\newblock Measure-valued derivatives for approximate bayesian inference.
\newblock \emph{4th workshop on Bayesian Deep Learning, NeurIPS}, 2019.

\bibitem[Ruiz et~al.(2016)Ruiz, Titsias, and Blei]{ruiz2016generalized}
Ruiz, F. J.~R., Titsias, M.~K., and Blei, D.~M.
\newblock The generalized reparameterization gradient.
\newblock In Lee, D.~D., Sugiyama, M., von Luxburg, U., Guyon, I., and Garnett,
  R. (eds.), \emph{Advances in Neural Information Processing Systems 29}, pp.\
  460--468, 2016.

\bibitem[Salimans \& Knowles(2014)Salimans and Knowles]{salimans2014using}
Salimans, T. and Knowles, D.~A.
\newblock On using control variates with stochastic approximation for
  variational bayes and its connection to stochastic linear regression.
\newblock \emph{arXiv preprint arXiv:1401.1022}, 2014.

\bibitem[Titsias \& L{\'a}zaro-Gredilla(2015)Titsias and
  L{\'a}zaro-Gredilla]{titsias2015local}
Titsias, M. and L{\'a}zaro-Gredilla, M.
\newblock Local expectation gradients for black box variational inference.
\newblock In \emph{Advances in neural information processing systems}, pp.\
  2620--2628. Citeseer, 2015.

\bibitem[Trivedi \& Zimmer(2007)Trivedi and Zimmer]{trivedi2007copula}
Trivedi, P.~K. and Zimmer, D.~M.
\newblock \emph{Copula modeling: an introduction for practitioners}.
\newblock ., 2007.

\bibitem[Tucker et~al.(2017)Tucker, Mnih, Maddison, Lawson, and
  Sohl{-}Dickstein]{tucker2017rebar}
Tucker, G., Mnih, A., Maddison, C.~J., Lawson, D., and Sohl{-}Dickstein, J.
\newblock {REBAR:} low-variance, unbiased gradient estimates for discrete
  latent variable models.
\newblock In \emph{Advances in Neural Information Processing Systems 30}, pp.\
  2627--2636, 2017.

\bibitem[Williams(1992)]{williams1992simple}
Williams, R.~J.
\newblock Simple statistical gradient-following algorithms for connectionist
  reinforcement learning.
\newblock \emph{Machine learning}, 8\penalty0 (3-4):\penalty0 229--256, 1992.

\bibitem[Wu et~al.(2019)Wu, Goodman, and Ermon]{wu2019differentiable}
Wu, M., Goodman, N., and Ermon, S.
\newblock Differentiable antithetic sampling for variance reduction in
  stochastic variational inference.
\newblock In \emph{The 22nd International Conference on Artificial Intelligence
  and Statistics}, pp.\  2877--2886. PMLR, 2019.

\bibitem[Yin \& Zhou(2019)Yin and Zhou]{yin2019arm}
Yin, M. and Zhou, M.
\newblock {ARM:} augment-reinforce-merge gradient for stochastic binary
  networks.
\newblock In \emph{7th International Conference on Learning Representations,
  {ICLR}}, 2019.

\bibitem[Yin et~al.(2019)Yin, Yue, and Zhou]{yin2019arsm}
Yin, M., Yue, Y., and Zhou, M.
\newblock {ARSM:} augment-reinforce-swap-merge estimator for gradient
  backpropagation through categorical variables.
\newblock In Chaudhuri, K. and Salakhutdinov, R. (eds.), \emph{Proceedings of
  the 36th International Conference on Machine Learning}, volume~97, pp.\
  7095--7104. {PMLR}, 2019.

\bibitem[{Yin} et~al.(2020){Yin}, {Ho}, {Yan}, {Qian}, and
  {Zhou}]{yin2020probabilistic}
{Yin}, M., {Ho}, N., {Yan}, B., {Qian}, X., and {Zhou}, M.
\newblock {Probabilistic Best Subset Selection by Gradient-Based Optimization}.
\newblock \emph{arXiv e-prints}, 2020.

\end{thebibliography}
\bibliographystyle{icml2021}

\onecolumn

\appendix
\centerline{\LARGE\textbf{Appendix}}

\section{Proofs}
\label{appendix:proofs}

\subsection*{Eq~\ref{eq:baseline}. LOORF unbiasedness and alternate form}
It is easy to see that for independent samples LOORF is unbiased, because $ \E [ \nabla_{\phi} \ln p(b) ] = 0 $. Using the linearity property of expectations and the fact that for independent variables $ \E [XY] = \E [X] \E [Y] $, we have:
\begin{align*}
    \E \Bigg[ \bigg( f(b_i) - \frac{1}{n-1} \sum_{j \neq i} f(b_j) \bigg) \nabla_{\phi} \ln p(b_i) \Bigg] &= \E \big[ f(b_i) \nabla_{\phi} \ln p(b_i) \big] - \frac{1}{n - 1}  \sum_{j \neq i} \E \big[ f(b_j) \big] \E[ \nabla_{\phi} \ln p(b_i) \big] \\
    &= \E \big[ f(b_i) \nabla_{\phi} \ln p(b_i) \big] = \nabla_{\bm \phi} \mathcal{E}(\bm \phi).
\end{align*}
The alternate form of LOORF is useful because we can easily calculate the mean and subtract it from each sample. For the $i^{\text{th}}$ sample, we have:
\begin{align*}
    &\Bigg( f(b_i) - \frac{1}{n-1} \sum_{j \neq i} f(b_j) \Bigg) \nabla_{\phi} \ln p(b_i) 
    = \frac{n}{n - 1} \Bigg( \frac{n - 1}{n} f(b_i) - \frac{1}{n} \sum_{j \neq i} f(b_j) \Bigg) \nabla_{\phi} \ln p(b_i) \\
    &= \frac{n}{n - 1} \Bigg( \Big(1 - \frac{1}{n} \Big) f(b_i) - \frac{1}{n} \sum_{j \neq i} f(b_j) \Bigg) \nabla_{\phi} \ln p(b_i) 
    = \frac{n}{n - 1} \Bigg( f(b_i) - \frac{1}{n} \sum_{i=1}^n f(b_j) \Bigg) \nabla_{\phi} \ln p(b_i).
\end{align*}

\subsection*{Eq~\ref{eq:pod}. PoD for binary variables}
First, we show that LOORF for $n=2$ samples is indeed PoD:
\begin{align*}
    \g{loorf}(b_1, b_2) &= \frac{1}{2} \bigg( \big( f(b_1) -  f(b_2) \big) (\nabla_{\phi} \ln p(b_1)) + \big( f(b_2) - f(b_1) \big) (\nabla_{\phi} \ln p(b_2)) \bigg) \\
    &= \frac{1}{2} \big( f(b_1) -  f(b_2) \big) \big( \nabla_{\phi} \ln p(b_1) - \nabla_{\phi} \ln p(b_2) \big) = \g{pod}(b_1, b_2).
\end{align*}
When both variables are Bern($p$), the score function is $\nabla_{\phi} \ln p_{\phi}(b) = b - \sigma(\phi)$, which simplifies the estimator:
\begin{align*}
    \g{pod}(b, b') = \big( f(b) -  f(b') \big) \big( b - \sigma(\phi) - b' + \sigma(\phi) \big) = \big( f(b) -  f(b') \big) \big( b - b' \big).
\end{align*}
For the expected value, first note that symmetric marginals $P(b = 1) = P(b' = 1)$ imply:
\begin{align*}
    P(b = 1, b' = 0) = P(b = 1) - P(b = 1, b' = 1) = P(b' = 1) - P(b = 1, b' = 1) = P(b = 0, b' = 1),
\end{align*}
and also that the estimator is zero when $b = b'$. With these two things in mind, we have:
\begin{align*}
    \E \big[ \g{pod}(b, b') \big] &= \frac{1}{2} P(b = 1, b' = 0) \big( f(1) - f(0) \big) (1 - 0) + \frac{1}{2} P(b = 0, b' = 1) \big( f(0) - f(1) \big) (0 - 1) \\
    &= \frac{1}{2} \big( P(b = 1, b' = 0) + P(b = 0, b' = 1) \big) \big( f(1) - f(0) \big) = \big( f(1) - f(0) \big) P(b = 1, b' = 0).
\end{align*}
If $b$ and $b'$ are independent $P(b = 1, b' = 0) = p(1 - p)$, which indeed coincides with the analytical gradient:
\begin{align*}
    \nabla_{\phi} \mathcal{E}(\phi) &= \nabla_{\phi} \E [ f(b) ] = \nabla_{\phi} \Big( \sigma(\phi) f(1) + (1 - \sigma(\phi)) f(0) \Big) =  \sigma(\phi)(1 - \sigma(\phi)) f(1) - \sigma(\phi)(1 - \sigma(\phi)) f(0) \\
    &= \big( f(1) - f(0) \big) \sigma(\phi)(1 - \sigma(\phi)) = \big( f(1) - f(0) \big) p(1 - p),
\end{align*}
where we used the fact that $\nabla_\phi \sigma(\phi) = \sigma(\phi)(1 - \sigma(\phi))$.

\subsection*{Theorem~\ref{thm:optimality}. Variance of ARTS}
The easiest way to compare the variance of ARTS to PoD is to just compute both. Define $\Delta f = f(1) - f(0)$, and the shorthand for the true gradient: $\nabla_{\phi} = \nabla_\phi \mathcal{E}(\phi) = \Delta f p (1 - p)$. Since ARTS reduces to PoD for $\rho=0$, and we know their expected values ($\nabla_{\phi}$), the variance is:
\begin{align*}
    &\E \big[ \g{arts}^2 \big] = \E \Bigg[ \bigg( \frac{1}{2} \big( f(b) - f(b') \big) (b - b') \bigg)^2 \bigg( \frac{2p(1 - p)}{P(b \neq b')} \bigg)^2 \Bigg] = \frac{1}{P(b \neq b')} \big( \Delta f p(1 - p) \big) ^ 2 = \frac{\nabla_{\phi} ^2 }{P(b \neq b')} \\
    &\implies \Var (\g{arts}^2) = \E \big[ \g{arts}^2 \big] - \E \big[ \g{arts} \big]^2 = \nabla^2 \bigg( \frac{1}{P(b \neq b')} - 1 \bigg).
\end{align*}
Rewriting this using the correlation will make the relationship clear:
\begin{align*}
    \frac{1}{1 - \rho} = \frac{2p(1 - p)}{P(b \neq b')} &\implies \frac{1}{P(b \neq b')} = \frac{1}{2p(1 - p)(1 - \rho)} \\
    \rho < 0 \implies 2p(1 - p)(1 - \rho) > 2p(1 - p) &\implies \frac{1}{2p(1 - p)(1 - \rho)} < \frac{1}{2p(1 - p)},
\end{align*}
which implies, for $\rho < 0$:
\begin{align*}
    \Var(\g{arts}^2) = \nabla^2 \bigg( \frac{1}{2p(1 - p)(1 - \rho)} - 1 \bigg) < \nabla^2 \bigg( \frac{1}{2p(1 - p)} - 1 \bigg) = \Var(\g{pod}^2).
\end{align*}
It is clear from above that the variance is an increasing (typo in original paper) function of $\rho$, and the lowest value it can achieve is the lower limit on a correlation of two Bernoulli variables. This depends on maximizing the probability $P(b = 1, b' = 0)$ as seen below, for which we have: $P(b = 1, b' = 0) \leq P(b = 1)$ as well as $P(b = 1, b' = 0) \leq P(b' = 0)$, which implies $P(b = 1, b' = 0) \leq \min(p, 1 - p)$. Putting it all together:
\begin{align*}
    \rho = \frac{P(b = 1, b' = 1) - p^2}{p(1 - p)} = \frac{p - P(b = 1, b' = 0) - p^2}{p(1 - p)} = \frac{p - \min(p, 1 - p) - p^2}{p(1 - p)} = - \min \bigg(\frac{p}{1 - p}, \frac{1 - p}{p} \bigg).
\end{align*}
The lower limit is achieved when $P(b = 1, b' = 0) = P(u < p, (1 - u) > p) = P(u < p , u < (1 - p) ) = \min (p, 1 - p)$, with $u \sim \text{Unif}(0, 1)$, which corresponds to the DisARM/U2G sampling method. Lastly, the debiasing term is:
\begin{align*}
    \frac{1}{1 - \rho} = \frac{p(1 - p)}{P(b = 1, b' = 0)} = \frac{p(1 - p)}{\min(p, 1 - p)} = \max(p, 1 - p)
\end{align*}

\subsection*{Eq~\ref{eq:podisloorf}. LOORF is all PoD pairs}
\begin{align*}
    \g{loorf}(&\bm b) = \frac{1}{n} \sum_{i=1}^n \left( f(\bm b_i) - \frac{1}{n-1} \sum_{j \neq i} f(\bm b_j) \right) \nabla_{\bm \phi} \ln p(\bm b_i) 
    = \frac{1}{n} \sum_{i=1}^n f(\bm b_i) \nabla_{\bm \phi} \ln p(\bm b_i) - \frac{1}{n(n - 1)} \sum_{i=1}^n \nabla_{\bm \phi} \ln p(\bm b_i) \sum_{j \neq i} f(\bm b_j) \\
    &= \frac{1}{n(n - 1)} \sum_{i \neq j} \Big( f(\bm b_i) \nabla_{\bm \phi} \ln p(\bm b_i) - f(\bm b_i) \nabla_{\bm \phi} \ln p(\bm b_j) \Big) 
    = \frac{1}{n (n - 1)} \sum_{i \neq j} \frac{1}{2} (f(\bm b_i) - f(\bm b_j)) (\nabla_{\bm \phi} \ln p(\bm b_i) - \nabla_{\bm \phi} \ln p(\bm b_j)) \\
    &= \frac{1}{n(n - 1)} \sum_{i \neq j} \g{pod}(b_i, b_j).
\end{align*}

\section{Theorem~\ref{thm:dirichlet_bernoulli}. Dirichlet Copula Derivation}
\label{appendix:dirichlet}

The two things we need to derive are the univariate and bivariate CDF. For both, we will make use of the Dirichlet aggregation property~\cite{ng2011dirichlet}: if $\bm d = (d_1, ..., d_n) \sim \text{Dir}(\bm \alpha)$, where $\bm \alpha = (\alpha_1, .., \alpha_n)$, then:
\begin{equation*}
    \label{eq:aggregation}
    (d_1, ..., d_i + d_j, ..., d_n) \sim \text{Dir}(\alpha_1, ..., \alpha_i + \alpha_j, ..., \alpha_n).
\end{equation*}
This is easily seen to be true from the Gamma sampling procedure:
\begin{align*}
    g_i \sim \text{Gamma}(\alpha_i, \theta), \; d_i = \frac{g_i}{\sum_j g_j} \;\implies\; (d_1, ..., d_n) \sim \text{Dir}(\alpha_1, ..., \alpha_n),
\end{align*}
because adding any two elements has no influence on the distribution of the others, and the fact that $g_i + g_j \sim \text{Gamma}(\alpha_i + \alpha_j, 1)$.

\subsection{Univariate CDF}
Using the aggregation property, we see that the marginal distribution of $d_i$ is:
\begin{equation*}
    (d_i, \sum_{j \neq i} d_j) \sim \text{Dir}(\alpha_i, \sum_{j \neq i} \alpha_j) = \text{Beta}(\alpha_i, \sum_{j \neq i} \alpha_j).
\end{equation*}
The CDF of the Beta distribution is the regularized incomplete Beta function $I_x(a, b)$, which has a simple closed form value when $a=1$ or $b=1$:
\begin{equation*}
   I_x(a, b) = \frac{B(x, a, b)}{B(a, b)} = \frac{\int_0^x t^{a - 1} (1 - t)^{b - 1} dt}{\int_0^1 t^{a - 1} (1 - t)^{b - 1} dt} ,\quad I_x(1, b) = 1 - (1 - x)^b, \quad I_x(a, 1) = x^a 
.\end{equation*}
Since $1 - d_i \sim \text{Beta}(\sum_{j \neq i} \alpha_j, \alpha_i)$, given a sample $\bm d \sim \text{Dir}(\bm 1_n)$ we can obtain two different copula samples using the probability integral transform from Eq.~\ref{eq:cdftransform}. If $\bm \alpha = \bm 1_n$, then the two marginal CDFs are $I_x(1, n - 1)$ and $I_x(n - 1, 1)$ for $d_i$ and $1 - d_i$, respectively. 

\subsection{Bivariate CDF}

The bivariate CDF is necessary for calculating $\rho = \text{Corr}(\tilde b_i, \tilde b_j)$ because:
\begin{align*}
    \rho = \frac{E[ b_i b_j] - p^2}{p(1 - p)} = \frac{P(b_i = 1, b_j = 1 - p^2}{p(1 - p)} = \frac{P(\tilde u_i < p, \tilde u_j < p) - p^2 }{p(1 - p)}.
\end{align*}
The transformation $\tilde u_i = 1 - (1 - d_i)^{n - 1}$ implies $d_i = 1 - (1 - \tilde u_i)^{\frac{1}{n - 1}}$, which makes the unknown term:
\begin{align*}
    P(\tilde{u}_i < p, \tilde{u}_j < p) = P(d_i < 1 - (1 - p)^{1 / (n - 1)}, d_j < 1 - (1 - p)^{1 / (n - 1)}),
\end{align*}
so the calculation reduces to finding the bivariate Dirichlet distribution of $(d_i, d_j)$. Using the aggregation property again, the density of $(d_i, d_j)$ is equivalent in both of these cases:
\begin{align*}
    (d_1, ..., d_i, ...,  d_j, ..., d_n) \sim \text{Dir}(\alpha_1, ..., \alpha_i, ..., \alpha_j, ..., \alpha_n) \; \iff \; (d_i, d_j, \sum_{k \neq i, j} d_k) \sim \text{Dir}(\alpha_i, \alpha_j, \sum_{k \neq i, j} \alpha_k).
\end{align*}
When $\bm \alpha = \bm 1_n$, the density of $\text{Dir}(1, 1, n - 2)$ is: $f(x, y) = (n - 1)(n - 2)(1 - x - y)^{n - 3} $. Note that $P(x < a, y < b) = P(x > a, y > b) - 1 + P(x < a) + P(y < b)$, so we can equivalently solve for the survival function $P(x > a, y > b)$. Let $a = b = q = 1 - (1 - p)^{1 / (n - 1)}$. To calculate the survival function $P(x > q, y > q)$ we need to be careful about the limits of integration. The conditions are:
\begin{equation*}
    q < x < 1,\quad q < y < 1,\quad y < 1 - x,\quad x < 1 - q \quad\implies\quad q < y < 1 - x, \quad q < x < 1 - q
.\end{equation*} 
The last condition implies $q < 1 - q \implies 1 - 2q > 0$. When this holds, the integral becomes:
\begin{align*}
    \int_p^{1 - q} &\int_p^{1 - x} (n - 1)(n - 2)(1 - x - y)^{n - 3} dy dx = (n-1)(n-2) \int_p^{1 - q} - \frac{(1 - x - y)^{n-2}}{n-2} \Bigg|_{y=q}^{1 - x} dx \\ 
    &= (n-1) \int_q^{1 - q} (1 - x - q)^{n-2} dx = -(n-1) \frac{(1 - x - q)^{n-1}}{n-1} \Bigg|_{x=q}^{1 - q} = (1 - 2q)^{n-1}.
\end{align*}
Therefore, $P(x > q, y > q) = \max(1 - 2q, 0)^{n-1}$, which implies $P(\tilde{u}'_i < p, \tilde{u}'_j < p) = \max(0, 2p^{1 / (n-1)} - 1)^{n-1}$. For the other copula uniforms $\tilde{\bm u} = 1 - \tilde{\bm u}'$, we have:
\begin{align*}
    P(u_i < p, u_j < p)
    &= P( 1 - \tilde u_i < p, 1 - \tilde u_j < p) 
    = P(\tilde u_i > 1 - p, \tilde u_j > 1 - p) \\
    &= 1 - P(\tilde u_i < 1 - p) - P(\tilde u_j < 1 - p) + P(\tilde u_i < 1 - p, \tilde u_j < 1 - p) \\
    &= 2p - 1 + \max(0, 2(1 - p)^{1 / (n-1)} - 1).
\end{align*}
Putting it all together:
\begin{align*}
    P(\tilde{u}_i < p, \tilde{u}_j < p) &= 2p - 1 + \max(0, 2(1 - p)^{1 / (n-1)} - 1)^{n - 1} \; &\implies \; \rho = \frac{\max(0, 2(1 - p)^{\frac{1}{n-1}} - 1)^{n - 1} - (1 - p)^2}{p(1 - p)} \\
    P(\tilde{u}_i' < p, \tilde{u}_j' < p) &= \max(0, 2p^{1 / (n-1)} - 1)^{n - 1} \; &\implies \; \rho' = \frac{\max(0, 2p^{1 / (n-1)} - 1)^{n - 1} - p^2 }{p(1 - p)}.
\end{align*}

\section{Dirichlet Copula Joint Density}

To use the Dirichlet copula it is not necessary to know the joint densities of $\tilde{\bm u}$ or $\tilde{\bm u}'$, but it is possible to derive either of them using the multivariate change of variables theorem $ p_U(u_1, \dots, u_n) = | \text{det}(J) | p_D(d_1, \dots, d_n) $, where $| \text{det}(J) |$ denotes the absolute value of the determinant of the Jacobian, whose elements are $J_{ij} = \partial d_i / \partial u_j$. 

For ease of notation denote the copula sample $\bm u = \tilde{\bm u}'$, which uses the elementwise transformation $u_i = f(d_i) = (1 - d_i)^{n - 1}$, with inverse $d_i = f^{-1}(u_i) = 1 - u_i^{1/(n - 1)}$. Since $d_i$ depends only on $u_i$ the determinant of the Jacobian is diagonal and thus has a simple form:
\begin{align*}
    | \det(J) | = \bigg| \prod_{i=1}^n \frac{\partial }{\partial u_i} \Big( 1 - u_i^{1 / (n - 1)} \Big) \bigg| = (n - 1)^{-n} \prod_{i=1}^n u_i^{\frac{1}{n - 1} - 1}.
\end{align*}
For the Dirichlet density simplex condition, we have:
\begin{align*}
    1 = \sum_{i=1}^n d_i = \sum_{i=1}^n (1 - u_i^{1 / (n - 1)}) = n - \sum_{i=1}^n u_i^{1 / (n - 1)} \; \implies \; \sum_{i=1}^n u_i^{1 / (n - 1)} = n - 1
\end{align*}
Putting it all together, the density is:
\begin{align*}
    p(u_1, ..., u_n) = \frac{(n - 1)!}{(n - 1)^n} \prod_{i=1}^n u_i^{\frac{1}{n - 1} - 1}, \quad \text{such that} \quad \sum_{i=1}^n u_i^{1 / (n - 1)} = n - 1.
\end{align*}

\pagebreak
\section{Additional Results}
\label{appendix:results}

\begin{table*}[h]
    \vskip -0.2in
    \caption{Test log likelihoods for ELBO optimized VAEs using different estimators. Results are reported on three datasets: Dynamic MNIST, Fashion MNIST, and Omniglot, and for 4, 6, 8, and 10 samples. Results are averaged over five runs, with the best performing methods in bold.}
    \label{tab:testresults}
    \vskip 0.15in
    \begin{center}
    \begin{small}
    \begin{sc}
    \begin{tabular}{ccc|rrrrr}
    \toprule
    \multicolumn{3}{c}{Samples} & \multicolumn{1}{c}{ARMS-D} &  \multicolumn{1}{c}{ARMS-N} & \multicolumn{1}{c}{LOORF} & \multicolumn{1}{c}{DisARM} & \multicolumn{1}{c}{RELAX} \\
    \midrule
    \multirow{8}{*}{\rotatebox[origin=c]{90}{\textbf{Dynamic MNIST}}} & \multirow{4}{*}{\rotatebox{90}{Linear}} & 4 &  \textbf{-111.57} ± \textbf{0.13} &  \textbf{-111.47} ± \textbf{0.16} &  -111.67 ± 0.04 &  -112.71 ± 0.07 &  -112.57 ± 0.34 \\
    &           & 6 &  \textbf{-110.47} ± \textbf{0.02} &   \textbf{-110.4} ± \textbf{0.12} &  \textbf{-110.42} ± \textbf{0.03} &  -111.58 ± 0.02 &  -110.94 ± 0.14 \\
    &           & 8 &  \textbf{-109.73} ± \textbf{0.06} &  -110.08 ± 0.06 &   -109.88 ± 0.07 &  -111.28 ± 0.16 &   -110.08 ± 0.10 \\
    &           & 10 &  \textbf{-109.52} ± \textbf{0.07} &  -109.61 ± 0.14 &  -109.61 ± 0.04 &  -110.56 ± 0.11 &  -109.64 ± 0.15 \\ \cmidrule{3-8}
    & \multirow{4}{*}{\rotatebox[origin=c]{90}{Nonlinr}} & 4 &  -100.02 ± 0.24 &  -100.05 ± 0.17 &   \textbf{-99.63 ± 0.07} &  -101.65 ± 0.32 &  -101.72 ± 0.18 \\
    &           & 6 &    -99.67 ± 0.20 &    \textbf{-98.90 ± 0.01} &   -99.21 ± 0.23 &  -100.42 ± 0.18 &   -100.31 ± 0.3 \\
    &           & 8 &   \textbf{-98.89} ± \textbf{0.36} &   \textbf{-98.86} ± \textbf{0.09} &   -99.22 ± 0.41 &   -99.82 ± 0.25 &   -99.91 ± 0.07 \\
    &           & 10 &   \textbf{-98.71} ± \textbf{0.22} &   \textbf{-98.31} ± \textbf{0.44} &   \textbf{-98.35} ± \textbf{0.54} &   -99.73 ± 0.27 &    -99.79 ± 0.20 \\
    \midrule
    \end{tabular}
    \begin{tabular}{ccc|rrrrr}
        \multirow{8}{*}{\rotatebox[origin=c]{90}{\textbf{Fashion MNIST}}} & \multirow{4}{*}{\rotatebox[origin=c]{90}{Linear}} & 4 &  \textbf{-254.65} ± \textbf{0.17} &  \textbf{-254.76} ± \textbf{0.05} &   -254.89 ± 0.07 &  -256.12 ± 0.04 &  -255.65 ± 0.08  \\
        &           & 6 &  -254.04 ± 0.22 &  \textbf{-253.78} ± \textbf{0.08} &  -254.14 ± 0.18 &  -255.11 ± 0.09 &   -254.40 ± 0.18 \\
        &           & 8 &  \textbf{-253.43} ± \textbf{0.13} &  \textbf{-253.24} ± \textbf{0.31} &  -253.56 ± 0.14 &  -254.73 ± 0.07 &   \textbf{-253.49} ± \textbf{0.10} \\
        &           & 10 &  -253.38 ± 0.04 &  \textbf{-253.19} ± \textbf{0.17} &  \textbf{-253.28} ± \textbf{0.01} &  -253.87 ± 0.29 &  \textbf{-253.28} ± \textbf{0.13} \\ \cmidrule{3-8}
        & \multirow{4}{*}{\rotatebox[origin=c]{90}{Nonlinr}} & 4 &  \textbf{-238.25} ± \textbf{0.46} &  \textbf{-238.36} ± \textbf{0.11} &  \textbf{-238.42} ± \textbf{0.06} &  -239.19 ± 0.09 &  -239.45 ± 0.15 \\
        &           & 6 &  \textbf{-238.08} ± \textbf{0.25} &  \textbf{-238.01} ± \textbf{0.15} &  -238.31 ± 0.22 &  -238.59 ± 0.11 &  -238.87 ± 0.39 \\
        &           & 8 &  -238.01 ± 0.14 &  \textbf{-237.71} ± \textbf{0.22} &  -237.98 ± 0.18 &  -238.25 ± 0.18 &   -238.25 ± 0.20 \\
        &           & 10 &  \textbf{-237.79} ± \textbf{0.16} &  \textbf{-237.94 }± \textbf{0.02} &  -238.19 ± 0.01 &  -238.24 ± 0.13 &  -238.02 ± 0.27 \\
    \midrule
    \end{tabular}
    \begin{tabular}{ccc|rrrrr}
        \multirow{8}{*}{\rotatebox[origin=c]{90}{\textbf{Omniglot}}} & \multirow{4}{*}{\rotatebox[origin=c]{90}{Linear}} & 4 &  \textbf{-118.61} ± \textbf{0.08} &  -118.73 ± 0.03 &  \textbf{-118.63} ± \textbf{0.09} &  -119.66 ± 0.26 &  -119.11 ± 0.06 \\
          &           & 6 &   \textbf{-118.00} ± \textbf{0.02} &  -118.03 ± 0.06 &  -118.12 ± 0.18 &  -118.87 ± 0.13 &  -118.24 ± 0.05 \\
          &           & 8 &   \textbf{-117.60} ± \textbf{0.05} &  -117.66 ± 0.12 &   -117.74 ± 0.10 &   -118.41 ± 0.10 &  -117.71 ± 0.02\\
          &           & 10 &  \textbf{-117.33} ± \textbf{0.14} &  \textbf{-117.39} ± \textbf{0.04} &  -117.51 ± 0.02 &  -118.06 ± 0.01 &   \textbf{-117.40} ± \textbf{0.05} \\ \cmidrule{3-8}  
        & \multirow{4}{*}{\rotatebox[origin=c]{90}{Nonlinr}} & 4 &  -116.14 ± 0.55 &  \textbf{-115.88} ± \textbf{0.15} &  -116.03 ± 0.39 &  -117.45 ± 0.24 &  -118.54 ± 0.51 \\
        &           & 6 &  \textbf{-115.33} ± \textbf{0.12} &  \textbf{-115.27} ± \textbf{0.24} &  \textbf{-115.05} ± \textbf{0.35} &  -116.45 ± 0.12 &  -118.05 ± 0.27 \\
        &           & 8 &   \textbf{-114.71} ± \textbf{0.10} &   -114.79 ± 0.05 &  \textbf{-114.52} ± \textbf{0.25} &   -115.80 ± 0.17 &  -118.12 ± 0.15 \\
        &           & 10 &  \textbf{-114.39} ± \textbf{0.16} &   \textbf{-114.52} ± \textbf{0.40} &   -114.60 ± 0.36 &  -115.15 ± 0.34 &   -118.50 ± 0.06 \\
    \bottomrule
    \end{tabular}
    \end{sc}
    \end{small}
    \end{center}
    \vskip -0.5in
\end{table*}

\begin{table}[h]
    \caption{Test log likelihoods for different estimators optimizing the multi sample bound, with the best performing methods in bold.}
    \label{tab:mstestresults}
    \vskip -0.25in
    \begin{center}
    \begin{small}
    \begin{sc}
    \setlength\tabcolsep{3.5pt}
    \resizebox{0.51\columnwidth}{!}{
    \begin{tabular}{ccc|rrr}
    \toprule
    \multicolumn{3}{c}{Samples} & \multicolumn{1}{c}{ARMS} &  \multicolumn{1}{c}{DisARM} & \multicolumn{1}{c}{VIMCO} \\
    \midrule
         \multirow{8}{*}{\rotatebox[origin=c]{90}{\textbf{Dynamic MNIST}}} & \multirow{4}{*}{\rotatebox{90}{Linear}} & 4 &  \textbf{-111.82} ± \textbf{0.09} &  \textbf{-111.87} ± \textbf{0.07} &  -112.52 ± 0.23 \\
         &           & 6 &  -\textbf{111.27} ± \textbf{0.01} &  \textbf{-111.15} ± \textbf{0.17} &  -111.66 ± 0.44 \\
         &           & 8 &  \textbf{-110.11} ± \textbf{0.18} &   -110.50 ± 0.11 &  -110.57 ± 0.31 \\
         &           & 10 &  \textbf{-109.74} ± \textbf{0.12} &  -109.89 ± 0.25 &  -110.17 ± 0.28 \\ \cmidrule{3-6}  
         & \multirow{4}{*}{\rotatebox[origin=c]{90}{Nonlinr}} & 4 &   \textbf{-98.96} ± \textbf{0.03} &    -99.30 ± 0.14 &   -98.98 ± 0.09 \\
         &           & 6 &   \textbf{-97.51} ± \textbf{0.27} &    -98.35 ± 0.20 &   \textbf{-97.65} ± \textbf{0.09} \\
         &           & 8 &    \textbf{-97.05} ± \textbf{0.20} &   \textbf{-97.23} ± \textbf{0.31} &   -97.33 ± 0.04 \\
         &           & 10 &   \textbf{-96.08} ± \textbf{0.22} &   -96.65 ± 0.11 &   -96.55 ± 0.02 \\
    \midrule
    \end{tabular}
    } \resizebox{0.51\columnwidth}{!}{
    \begin{tabular}{ccc|rrr}
         \multirow{8}{*}{\rotatebox[origin=c]{90}{\textbf{Fashion MNIST}}} & \multirow{4}{*}{\rotatebox[origin=c]{90}{Linear}} & 4 &   \textbf{-254.34} ± \textbf{0.10} &   -254.48 ± 0.20 &  -254.86 ± 0.13 \\
         &           & 6 &  \textbf{-253.51} ± \textbf{0.07} &  \textbf{-253.41} ± \textbf{0.18} &  -253.76 ± 0.32 \\
         &           & 8 &  \textbf{-252.78} ± \textbf{0.13} &  \textbf{-252.72} ± \textbf{0.16} &   -252.90 ± 0.09 \\
         &           & 10 &  -252.58 ± 0.07 &    \textbf{-251.90} ± \textbf{0.10} &   -252.70 ± 0.15 \\ \cmidrule{3-6}  
         & \multirow{4}{*}{\rotatebox[origin=c]{90}{Nonlinr}} & 4 &  \textbf{-237.31} ± \textbf{0.18} &   -238.16 ± 0.10 &  -\textbf{237.28} ± \textbf{0.25} \\
         &           & 6 &  \textbf{-236.22} ± \textbf{0.52} &  \textbf{-236.62} ± \textbf{0.25} &  \textbf{-236.33} ± \textbf{0.58} \\
         &           & 8 &  -235.64 ± 0.02 &  \textbf{-236.37} ± \textbf{0.04} &   -235.96 ± 0.08 \\
         &           & 10 &  \textbf{-235.27} ± \textbf{0.04} &  -235.65 ± 0.27 &  -235.72 ± 0.06 \\
    \midrule
    \end{tabular}
    } \resizebox{0.51\columnwidth}{!}{
    \begin{tabular}{ccc|rrr}
      \multirow{8}{*}{\rotatebox[origin=c]{90}{\textbf{Omniglot}}} & \multirow{4}{*}{\rotatebox[origin=c]{90}{Linear}} & 4 &  \textbf{-119.56} ± \textbf{0.09} &  \textbf{-119.28} ± \textbf{0.31} &  -120.71 ± 0.16 \\
         &           & 6 &  -119.34 ± 0.37 &  \textbf{-119.09} ± \textbf{0.08} &  -119.65 ± 0.09 \\
         &           & 8 &   \textbf{-119.00} ± \textbf{0.11} &   \textbf{-118.90} ± \textbf{0.16} &  -119.08 ± 0.21 \\
         &           & 10 &   -118.82 ± 0.10 &  \textbf{-118.57} ± \textbf{0.08} &  -118.93 ± 0.17 \\ \cmidrule{3-6}  
         & \multirow{4}{*}{\rotatebox[origin=c]{90}{Nonlinr}} & 4 &  \textbf{-115.79} ± \textbf{0.35} &   -116.91 ± 0.40 &  \textbf{-115.87} ± \textbf{0.35} \\
         &           & 6 &  \textbf{-114.56} ± \textbf{0.25} &  -115.31 ± 0.42 &   \textbf{-114.68} ± \textbf{0.10} \\
         &           & 8 &  \textbf{-114.09} ± \textbf{0.09} &   -114.45 ± 0.30 &   \textbf{-113.89} ± \textbf{0.22} \\
         &           & 10 &   -113.40 ± 0.05 &  \textbf{-114.19} ± \textbf{0.16} &  -113.55 ± 0.05 \\
    \bottomrule
    \end{tabular}
    }
    \end{sc}
    \end{small}
    \end{center}
    \vskip -0.2in
\end{table}

\begin{figure*}[h]
    \begin{center}
        \centerline{\includegraphics[width=0.95\columnwidth]{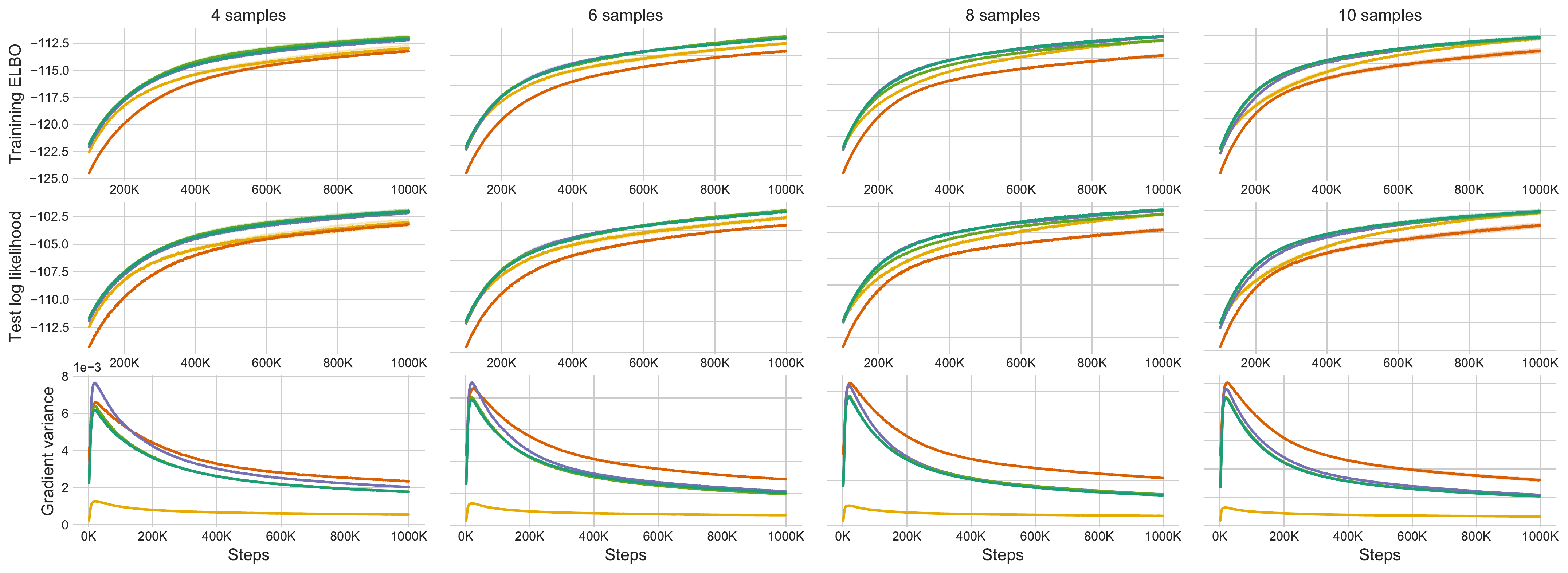}}
        \centerline{\includegraphics[width=0.95\columnwidth]{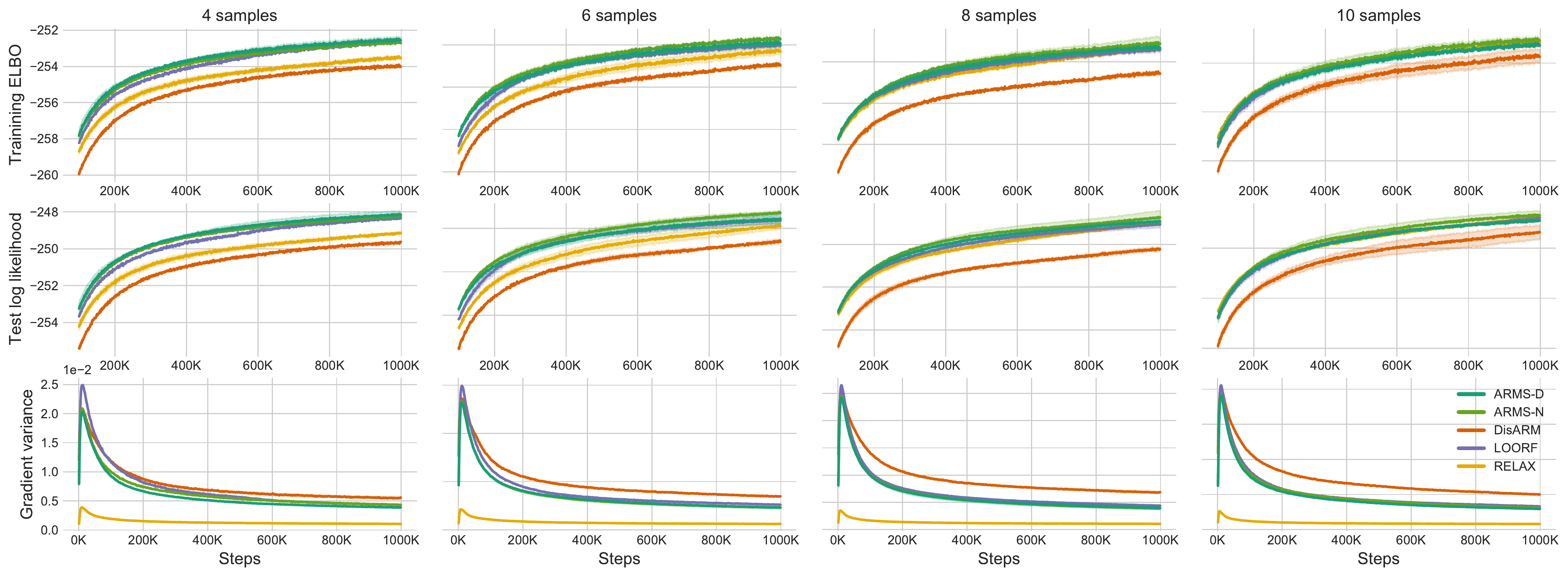}}
        \centerline{\includegraphics[width=0.95\columnwidth]{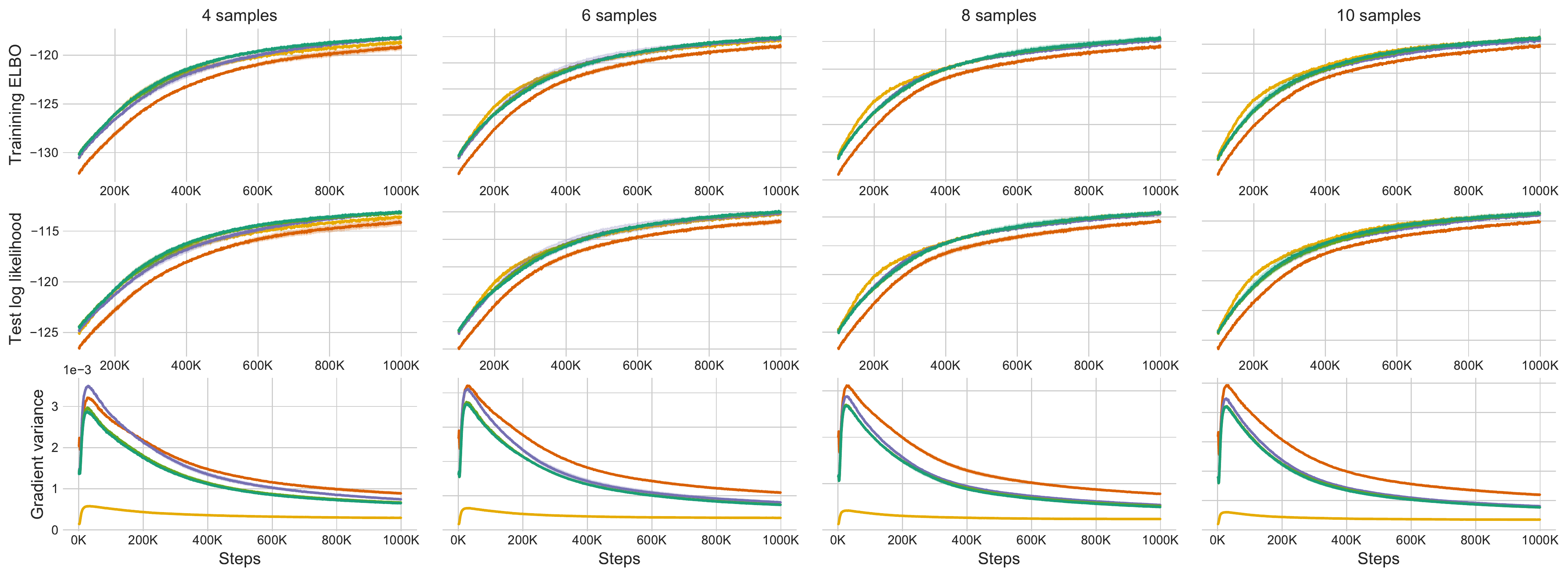}}
       \vspace{-4.5mm}
        \caption{For each dataset, shown is training a \textit{linear} discrete VAE using the \textit{ELBO}. Each group of three rows, from top to bottom, represent Dynamic MNIST, Fashion MNIST, and Omniglot, respectively. Within each triplet of rows, they correspond to the training ELBO, test log likelihood, and the variance of the gradient updates averaged over all parameters. Columns correspond to $n \in \{4, 6, 8, 10 \}$ samples used per step.}
    \end{center}
    \vspace{-4.5mm}
\end{figure*}

\begin{figure*}[h]
    \begin{center}
        \centerline{\includegraphics[width=0.95\columnwidth]{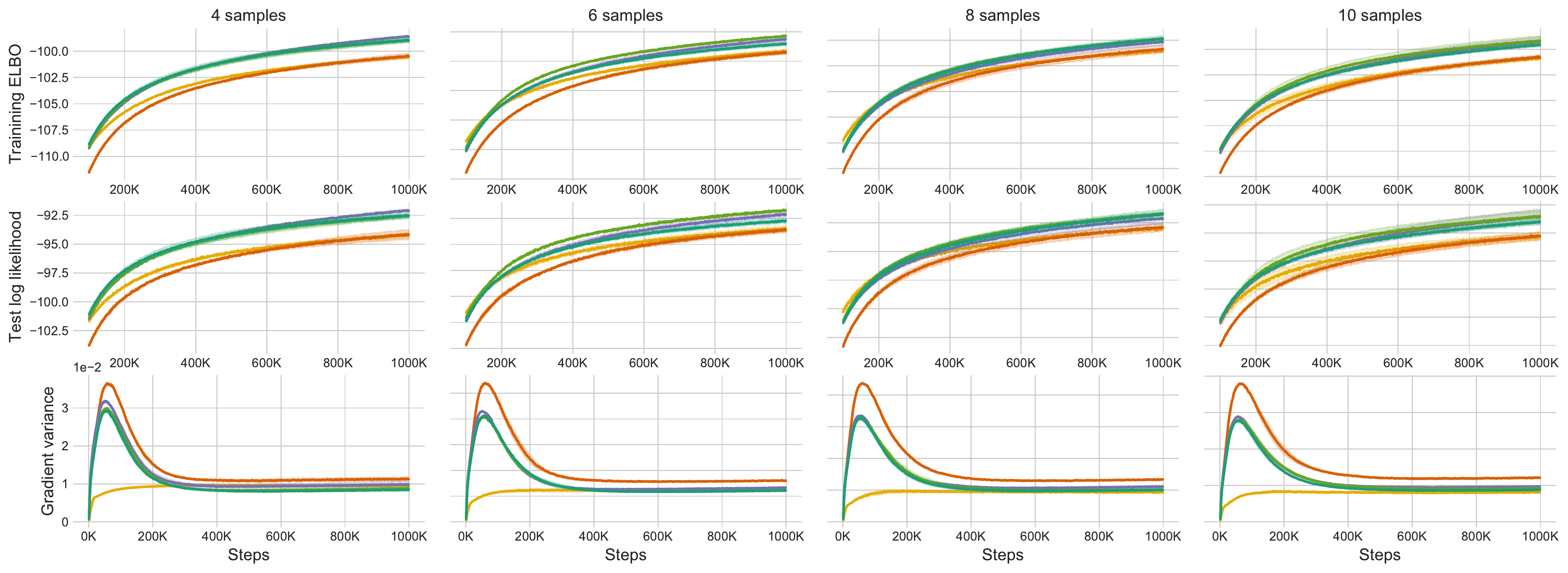}}
        \centerline{\includegraphics[width=0.95\columnwidth]{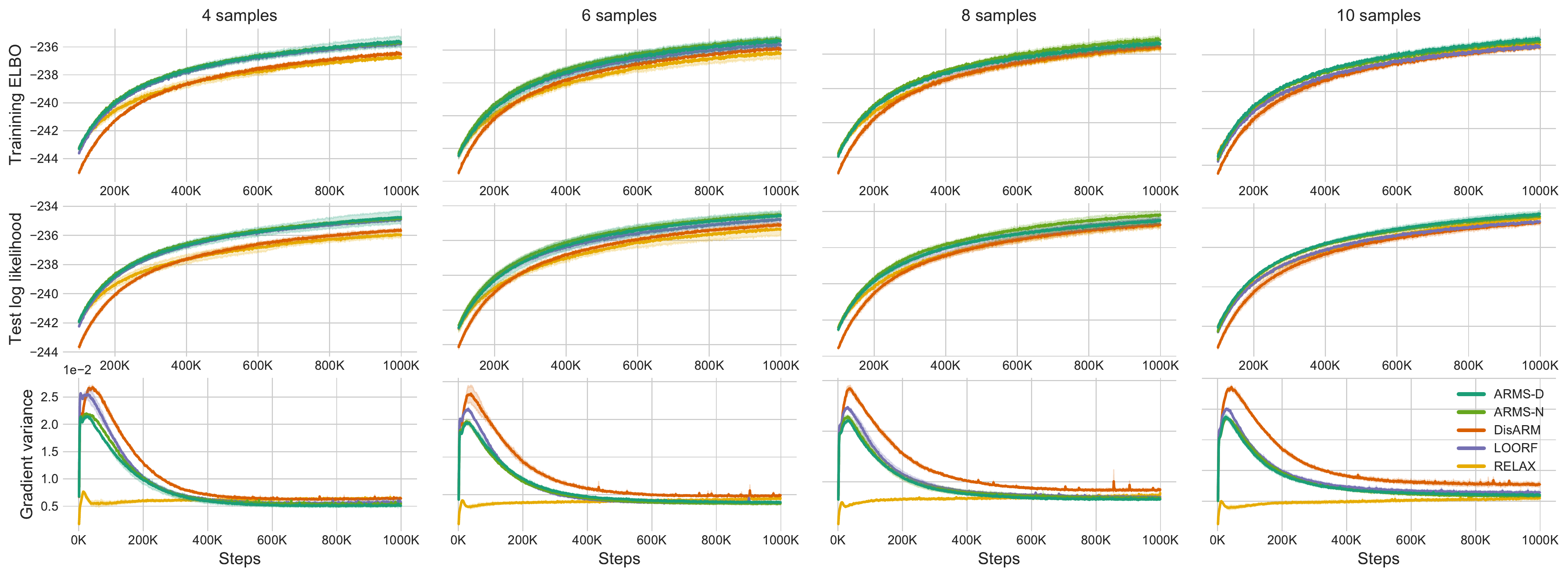}}
        \centerline{\includegraphics[width=0.95\columnwidth]{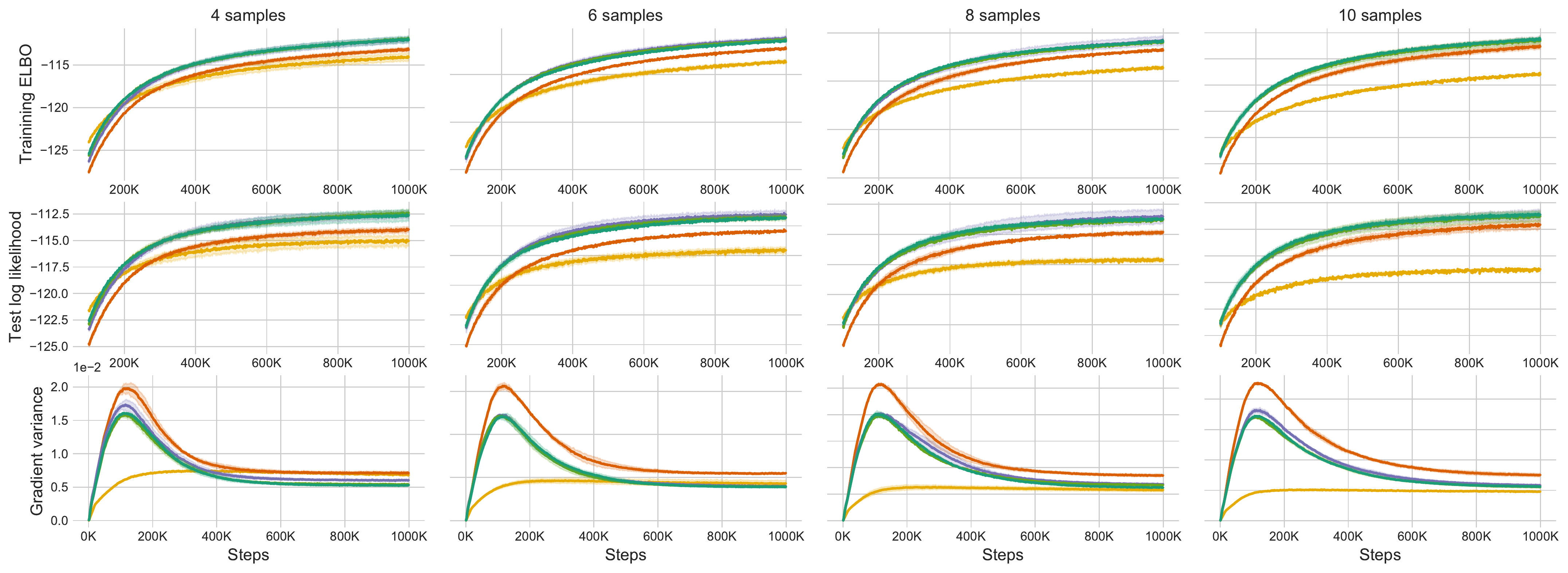}}
       \vspace{-4.5mm}
        \caption{For each dataset, shown is training a \textit{nonlinear} discrete VAE using the \textit{ELBO}. Each group of three rows, from top to bottom, represent Dynamic MNIST, Fashion MNIST, and Omniglot, respectively. Within each triplet of rows, they correspond to the training ELBO, test log likelihood, and the variance of the gradient updates averaged over all parameters. Columns correspond to $n \in \{4, 6, 8, 10 \}$ samples used per step.}
    \end{center}
    \vspace{-4.5mm}
\end{figure*}

\begin{figure*}[h]
    \begin{center}
        \centerline{\includegraphics[width=0.95\columnwidth]{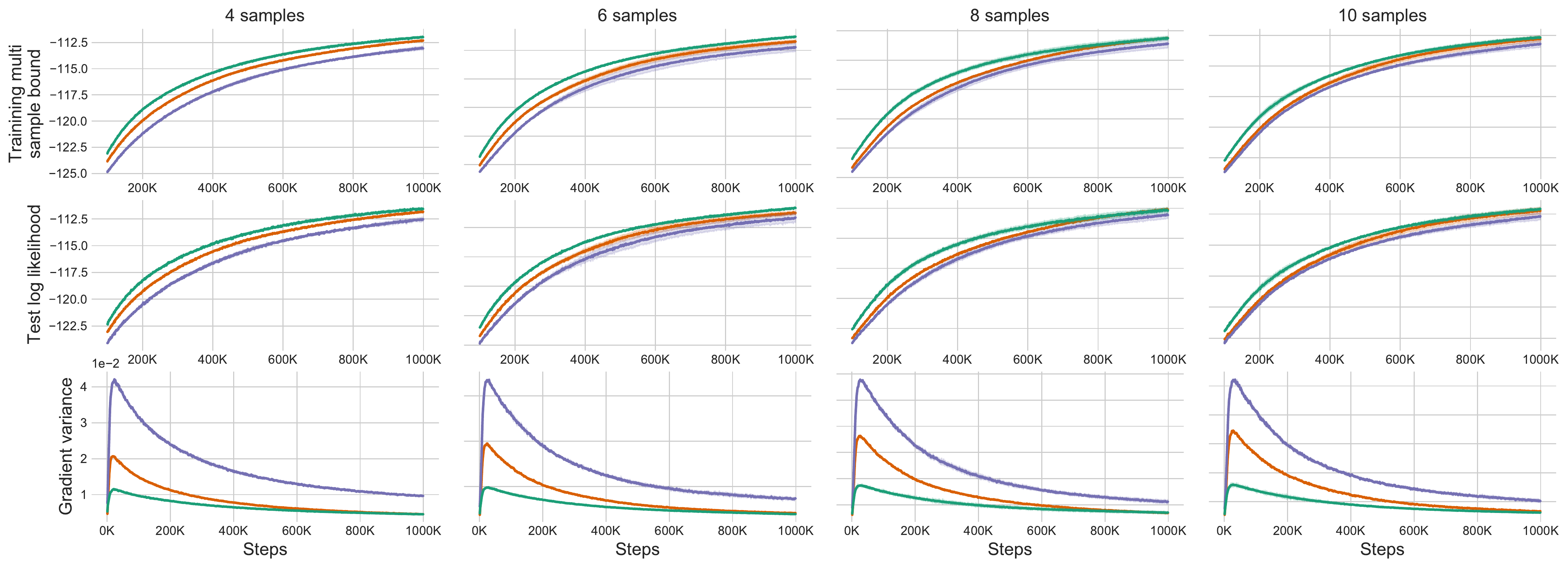}}
        \centerline{\includegraphics[width=0.95\columnwidth]{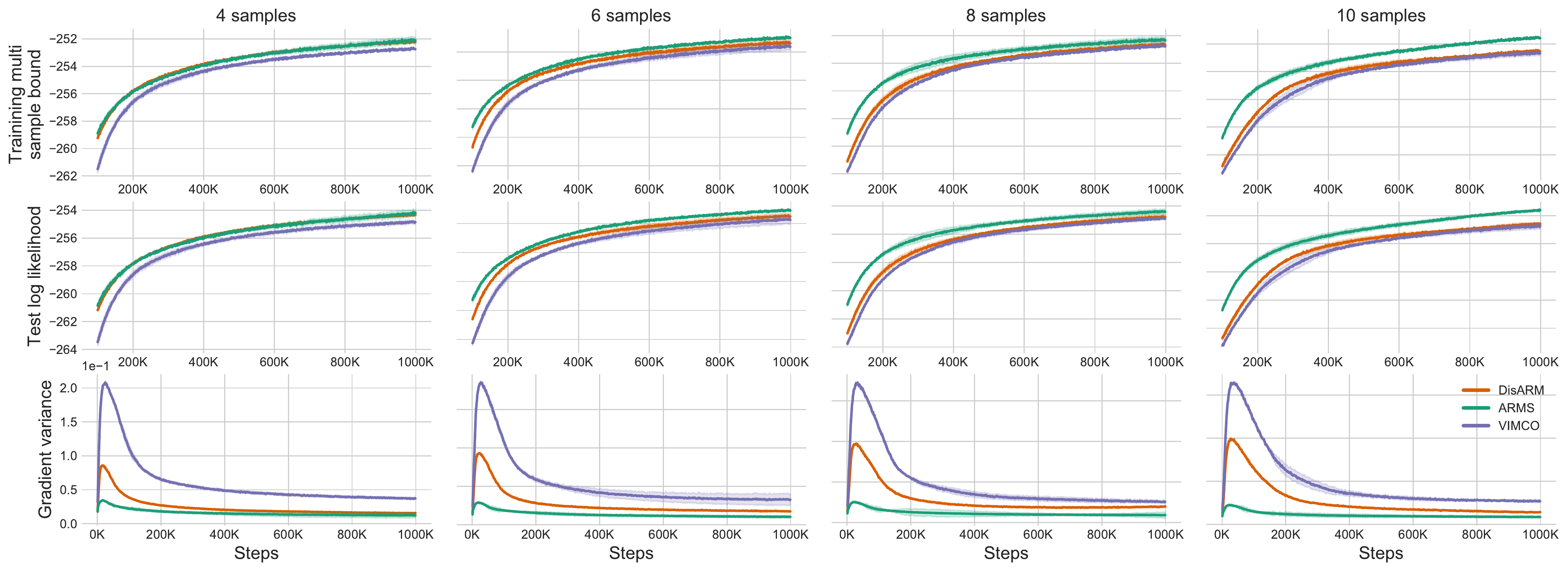}}
        \centerline{\includegraphics[width=0.95\columnwidth]{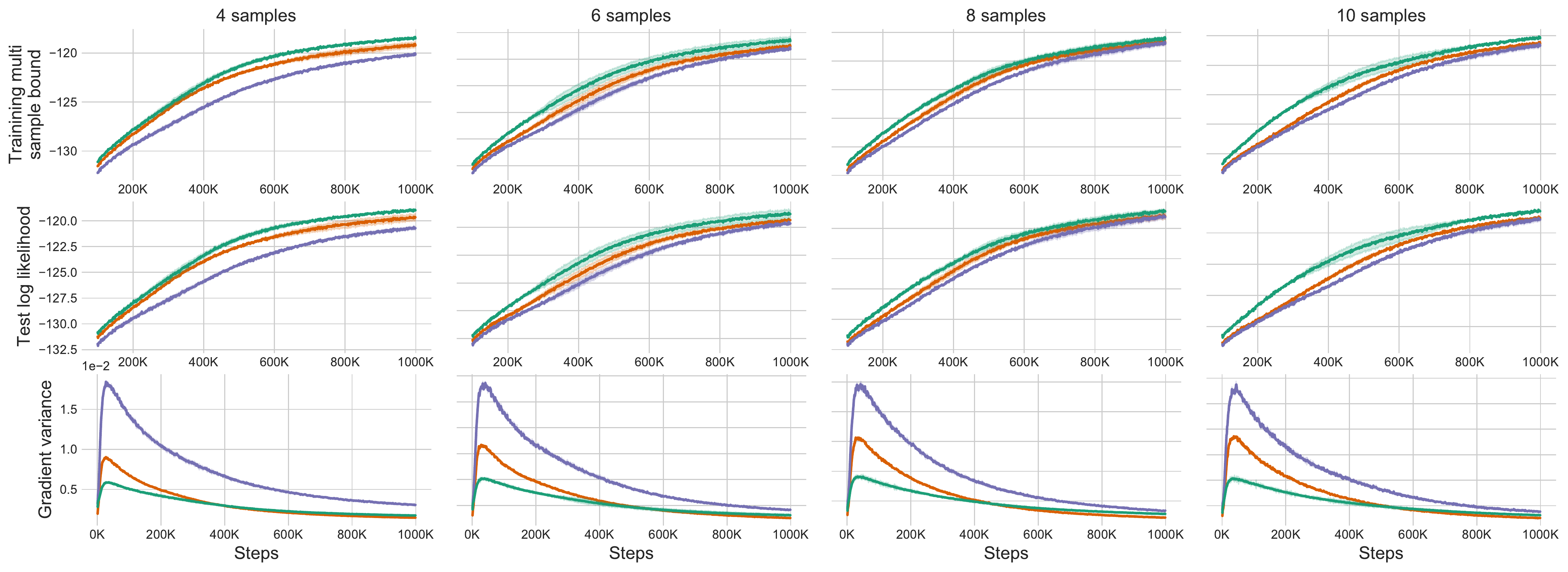}}
       \vspace{-4.5mm}
        \caption{For each dataset, shown is training a \textit{linear} discrete VAE using the \textit{multi sample bound}. Each group of three rows, from top to bottom, represent Dynamic MNIST, Fashion MNIST, and Omniglot, respectively. Within each triplet of rows, they correspond to the training multi sample bound, test log likelihood, and the variance of the gradient updates averaged over all parameters. Columns correspond to $n \in \{4, 6, 8, 10 \}$ samples used per step.}
    \end{center}
    \vspace{-4.5mm}
\end{figure*}

\begin{figure*}[h]
    \begin{center}
        \centerline{\includegraphics[width=0.95\columnwidth]{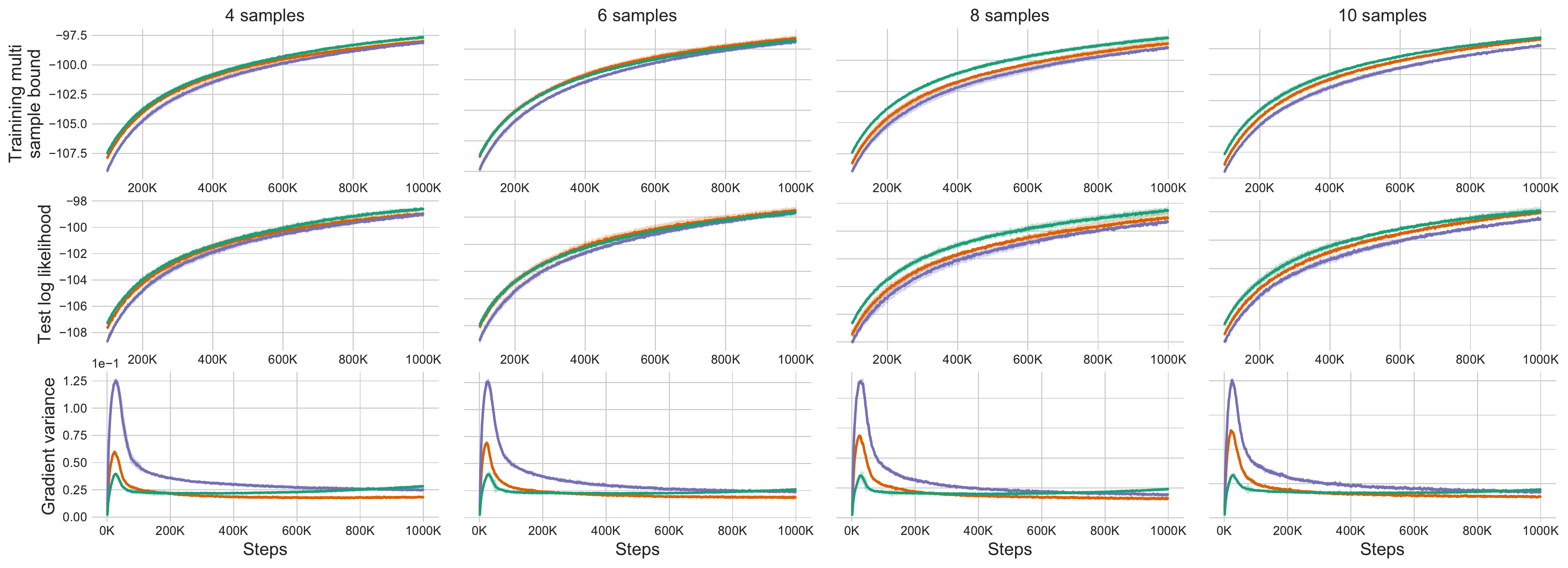}}
        \centerline{\includegraphics[width=0.95\columnwidth]{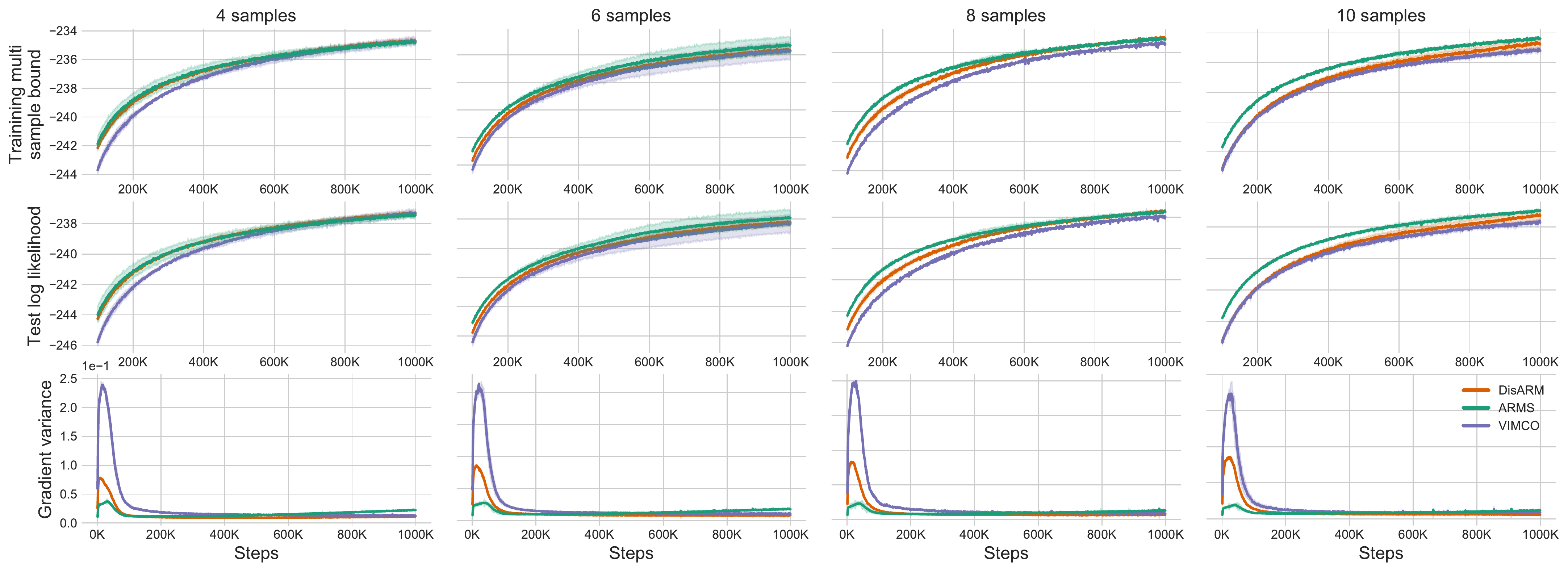}}
        \centerline{\includegraphics[width=0.95\columnwidth]{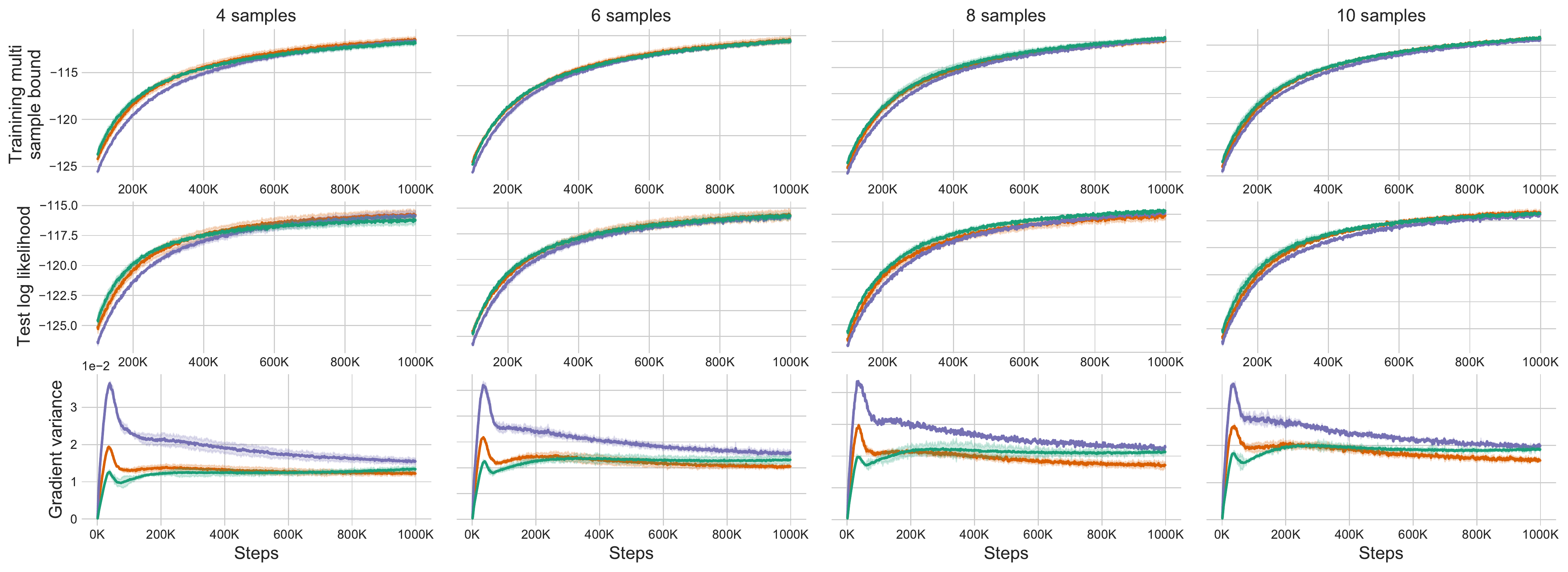}}
       \vspace{-4.5mm}
        \caption{For each dataset, shown is training a \textit{nonlinear} discrete VAE using the \textit{multi sample bound}. Each group of three rows, from top to bottom, represent Dynamic MNIST, Fashion MNIST, and Omniglot, respectively. Within each triplet of rows, they correspond to the training multi sample bound, test log likelihood, and the variance of the gradient updates averaged over all parameters. Columns correspond to $n \in \{4, 6, 8, 10 \}$ samples used per step.}
    \end{center}
    \vspace{-4.5mm}
\end{figure*}

\end{document}